\documentclass[journal]{IEEEtran}
\usepackage{cite}
\usepackage{times}
\usepackage{epsfig}
\usepackage{graphicx}
\usepackage{amsmath}
\usepackage{amssymb}
\usepackage{times}  
\usepackage{helvet}  
\usepackage{courier}  
\usepackage{url}  
\usepackage{graphicx}  
\usepackage{multirow}
\usepackage{tabularx} 
\usepackage{caption}
\usepackage{xspace}
\usepackage{booktabs}
\usepackage{subfigure}
\usepackage{amsmath,amssymb} 
\usepackage[ruled,vlined]{algorithm2e}
\usepackage{textcomp}

\ifCLASSINFOpdf
\else
\fi
\begin{document}
%
\title{Recurrent Distillation based Crowd Counting}
%
%
%

\author{Yue Gu,
        Wenxi Liu 
\thanks{Y. Gu and W. Liu was with the College of Mathematics and Computer Science, Fuzhou University}}

\maketitle

\begin{abstract}
In recent years, with the progress of deep learning technologies, crowd counting has been rapidly developed. In this work, we propose a simple yet effective crowd counting framework that is able to achieve the state-of-the-art performance on various crowded scenes. In particular, we first introduce a perspective-aware density map generation method that is able to produce ground-truth density maps from point annotations to train crowd counting model to accomplish superior performance than prior density map generation techniques. Besides, leveraging our density map generation method, we propose an iterative distillation algorithm to progressively enhance our model with identical network structures, without significantly sacrificing the dimension of the output density maps. In experiments, we demonstrate that, with our simple convolutional neural network architecture strengthened by our proposed training algorithm, our model is able to outperform or be comparable with the state-of-the-art methods. Furthermore, we also evaluate our density map generation approach and distillation algorithm in ablation studies.
\end{abstract}


%
\IEEEpeerreviewmaketitle

\section{Introduction}
\label{sec:introduction}
In recent years, vision-based crowd analysis has been extensively researched, due to its wide applications in crowd management, traffic control, urban planning, and surveillance. As one of the most important applications, crowd counting has been studied extensively \cite{Zhang_2016_CVPR_MCNN,Idrees_2018_ECCV_CL,Ma_2019_ICCV_BL}. With the recent progress of deep learning techniques, the performance of crowd counting has been significantly elevated \cite{Li_2018_CVPR_CSRNet,Ma_2019_ICCV_BL,Yan_2019_ICCV_PGCNet,Liu_2019_ICCV_DSSINet}. The convolutional neural network (CNN)-based methods have demonstrated excellent performance on the task of counting dense crowds in images. 
Most of CNN-based methods first estimate the density map via deep neural networks and then calculate the counts \cite{Zhang_2016_CVPR_MCNN,Li_2018_CVPR_CSRNet,Cao_2018_ECCV_SA,Wan_2019_ICCV_adaptive_map,Yan_2019_ICCV_PGCNet}. In specific, the concept of {density map}, where the integral (sum) over any sub-region equals the number of objects in that region, was first proposed in \cite{NIPS2010_4043_map}. 
Since the existing crowd counting benchmarks provide the point annotation for each crowd image, in which each point is located on the head of a person in the crowd. To train a CNN-based crowd counting model, the point annotations need to be converted to a density map in advance.
Lempitsky \textit{et al.} \cite{NIPS2010_4043_map} propose to use a normalized 2D Gaussian kernel to convert the point annotations to a ground-truth density map. Typically, the Gaussian kernel size is fixed while converting point annotations. But this trivial approach degrades the counting performance, since the scales of individuals in the crowd image may vary greatly. To produce better ground-truth density map, Zhang \textit{et al.} \cite{zhang2015cross} provides a manual estimation of the perspective maps for the crowd images. But it is laborious to provide the accurate perspective information for all the image captured in various scenarios. Zhang \textit{et al.} \cite{Zhang_2016_CVPR_MCNN} introduce the geometry-adaptive kernel to create ground-truth density maps. They assume that the crowd is evenly distributed and thus they can estimate the kernel size by the average distance between each point annotations and its nearest neighbors.
Generally, geometry-adaptive kernels is efficient for estimating the kernel of dense point annotations, but it is inaccurate when the crowd is not distributed evenly.
As we know, with different Gaussian kernel sizes, the ground-truth density maps converted from the point annotations can be sharper with smaller kernel sizes or smoother with larger kernel sizes (as shown in Fig.~\ref{fig:density_map}(b) and (c)). Empirically, it is easier for a CNN model to fit the smoother crowd density maps rather than the sharper ones. This is probably because the sharper ground-truth density maps contain a larger amount of zero values than the non-zero ones, which makes the network difficult to fit the ground-truth.
Nevertheless, the individual information contained in the smoother density maps is relatively vague compared with the sharper ones. So the performance of the crowd counting model will thus be degraded and the prior density map generation approaches can hardly handle this problem. 

To mitigate this problem, our work attempts to tackle it in several aspects. 
First, we propose a parametric perspective-aware density map generation approach. We assume that the individuals on the same horizontal line (or the same row) of the image are from the similar distance away from the camera, so the kernel sizes for the point annotations on the same row should be the same. 
Thus, to determine the kernel size for each row of the image, we introduce a metric, \textit{effective density}, used to measure the density of the highly aggregated segments on each row. By linearly mapping the effective densities to the manually-defined kernel size range, we can easily produce perspective-aware density maps by assigning a small number of parameters as the ground-truths of model training.
Second, we propose a simple CNN-based architecture featuring with two output branches which are supervised by a multi-task loss with low-resolution and high-resolution ground-truth density maps. With the supervision of high- and low-resolution density maps, our model is able to generate the crowd distribution with the relatively high dimension (i.e., $1/4$ of the input size). 

Last but not least, we propose an iterative distillation optimization algorithm for progressively enhancing the performance of our network model. As mentioned, although our network can regress the crowd distribution, its performance is still constrained by the quality of the density maps. To benefit learning from more accurate yet hard-to-learn density maps (i.e., the sharp density map generated by small Gaussian kernel), we propose to iteratively distill the network with the previously trained identical network. As known, the distillation techniques have been previously used to compress network or improve network capability \cite{hinton2015distilling,yang2019snapshot,furlanello2018born}. Here, we employ it to strengthen the capability of our network, when the training objective of the crowd counting model becomes more and more challenging. Particularly, during distillation, our proposed parametric density map generation approach can be iteratively utilized to generate sharper ground-truth density maps as the training objectives. In experiments, we demonstrate that our perspective-aware density map generation method is better than the prior generation techniques. Besides, we show that, although our network architecture is simple, our approach can still obtain the state-of-the-art performance compared with other methods.

In summary, our contributions are below:
\begin{itemize}
	\item We introduce a parameteric perspective-aware density map generation method to generate ground-truth density maps from crowd point annotations, so as to train a crowd counting model that can estimate the crowd density maps with relatively high spatial dimension.
	\item We present a novel iterative distillation algorithm to enhance our model while progressively reducing the Gaussian kernel sizes of the ground-truth density maps, which can further improve the counting performance.
	\item In experiments, we show that our simple network architecture can reach the state-of-the-art performance compared against the latest approaches in public benchmarks. 
\end{itemize}

In the following, we first review the related works in Sec.~\ref{sec:rw}. Then, we elaborate the methodology in Sec.~\ref{sec:method}. Lastly, in Sec.~\ref{sec:exp}, we evaluate our proposed approach in public benchmarks.

\section{Related work}
\label{sec:rw}

In this section, we survey the most related works on density map generation and knowledge distillation.

\subsection{Density map generation for crowd counting}

The perspective distortion in images is one of the main factors that affect the accurate counting of the crowd. Due to the shooting scene, angle, terrain distribution and other factors, the perspective of each picture is different. It is a natural idea to combine perspective information to improve the ground-truth density map. By manually measuring the height of people in different positions in the picture, \cite{zhang2015cross} calculated the perspective of 108 scenes in WorldExpo'10 dataset , and used the perspective to generate an appropriate ground\_truth density map for the dataset.
However, most datasets contain various scenes (such as ShangHaiTech~\cite{Zhang_2016_CVPR_MCNN}, UCF\_QNRF~\cite{Idrees_2018_ECCV_CL}), so manually estimating the perspective of each image is very laborious. Thus, Zhang \textit{et al.} \cite{Zhang_2016_CVPR_MCNN} proposed the adaptive geometric estimation technique. The Gaussian covariance of each marker point is estimated by calculating its average distance from the surrounding points, thereby generating a ground-truth density map that more closely matches the actual distribution of the crowd.
Shi \textit{et al.} \cite{shi2019counting} proposed a non-uniform kernel estimation. They assumes that the crowd is unevenly distributed throughout the crowd image. 
They first used {adaptive geometric estimation technique} to estimate the covariance of all points, 
and then calculated the average covariance of the local neighborhood of each point, which can thus make the Gaussian distribution of a single point will not become too large or too small.
Compared with these methods, we propose a new method to produce a ground-truth density map, by assuming the points on each row of the image have the Gaussian kernels with the similar size and estimating the perspective of the entire image via rough density maps without any supervision.

\subsection{Knowledge distillation}
Knowlege distillation is a deep learning training technique that trains a network model (i.e., the student model) to mimic the behavior of a independently trained model (i.e., the teacher model). It was originally used for model compression when the student model is light-weighted. However, some recent studies~\cite{furlanello2018born,yang2019snapshot} have found that the distillation between the models with the identical network structure can obtain better classification results. 
Particularly, Yang \textit{et al.} \cite{yang2019snapshot} proposed the snapshot distillation method, which divides the training process into two stages. In the first stage, the network performs normal training. Then use the first-stage model as a teacher model to guide the student model for the second-stage training. Inspired by these works, we propose our distillation algorithm for the task of crowd counting.

\section{Our Proposed Method}
\label{sec:method}

In this section, we first introduce our deep network architecture. To progressively strengthen the performance of our network, we propose a distillation based optimization approach as well as our parametric density map generation method.

\subsection{Network architecture}

In order to train a crowd counting model, we follow the principle of the prior frameworks (e.g.~\cite{zhang2015cross}), in which the deep neural network model aims to generate a density map from the input crowd image and its crowd count can be measured by accumulating the values of the entire density map.

\begin{figure}
	\centering
	\includegraphics[width=1\linewidth]{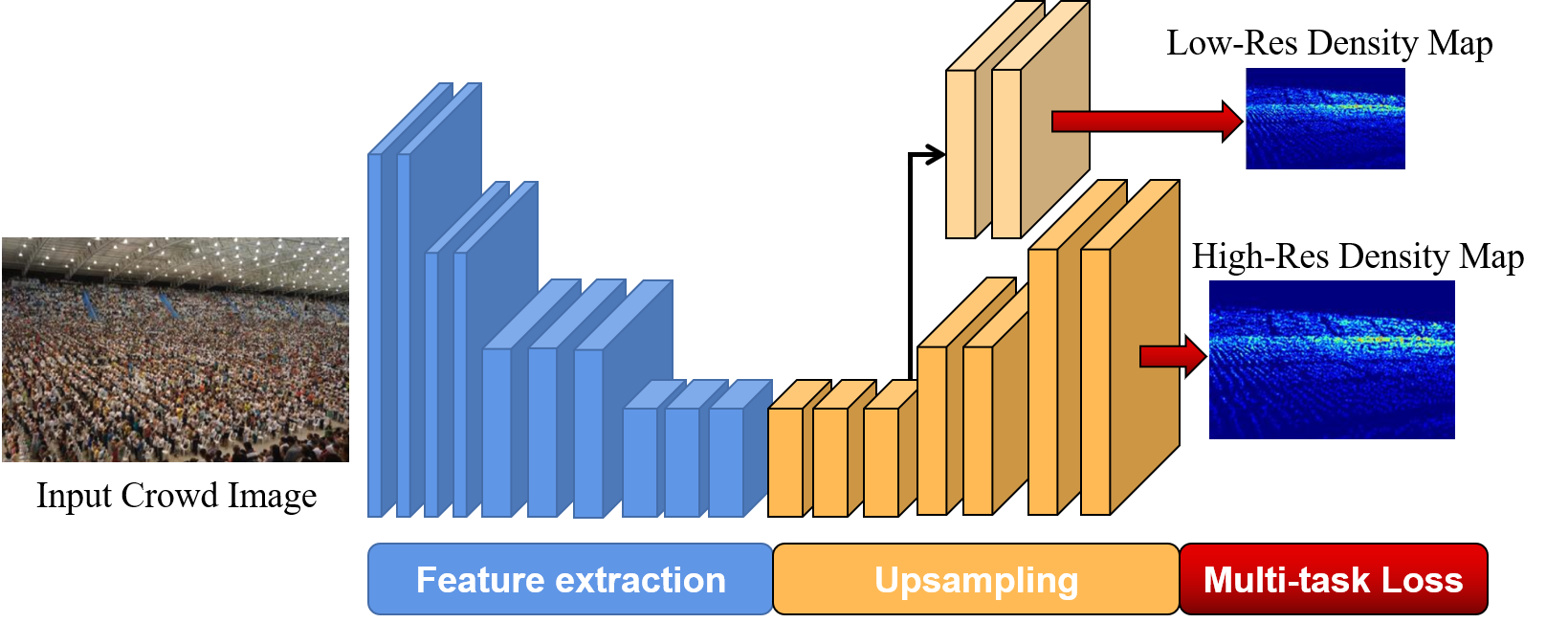}
	\caption{Illustration of our crowd counting network model. As shown, our framework is composed of feature extraction, upsampling, and multi-task loss. Particularly, the feature representation of the input crowd image can be extracted via pretrained convolutional layers. Through upsampling layers, our network reconstructs the extracted features to density maps. We introduce a multi-task loss to simultaneously supervise the high-resolution and low-resolution density map generation. 
		The high resolution ground-truth is $1/4$ of the input image, while the low resolution one is $1/16$ of the input image. }
	\label{fig:network}
\end{figure}

As shown in Fig.~\ref{fig:network}, we adopt a vanilla deep convolutional neural network (CNN) as the backbone network to estimate the count of a crowd image, which consists of a downsampling module and an upsampling module. The downsampling module adopts the first 10 convolutional layers of the pretrained model VGG-16 \cite{simonyan2014very}, which extracts deep visual feature representation from the input crowd image. Then, two transposed convolutional layers are applied to upsample the spatial dimension of the feature maps to $1/4$ of the input dimension. As depicted in Table~\ref{tab:resolution}, most prior counting models produce low-resolution density map (i.e., less than 1/16 of the original size), yet producing a high-resolution density map benefits many downstream applications such as crowd analysis. 
The reason we adopt such a simple network architecture is to demonstrate the performance of our proposed density map generation and distillation method.

\begin{table}[t]
	\setlength{\tabcolsep}{6pt}
	\centering
	\caption{The ratio of the dimensions of the output density maps and the input image for different crowd counting methods. }
	\begin{tabular}{c|c|c}
		\toprule
		Method         & Year        & Ratio  \\ 
		\midrule
		MBTTBF-SCFB \cite{Sindagi_2019_ICCV_MBTTBF}     & 2019 &1/256  \\
		CSRNet \cite{Li_2018_CVPR_CSRNet}                  &2018  &1/64  \\ 
		ADCrowdNet \cite{Liu_2019_CVPR_ADCrowd}  & 2019 &1/64  \\ 
		CSRNet+PACNN \cite{Shi_2019_CVPR_PACNN} & 2019&1/64  \\ 
		CAN \cite{Liu_2019_CVPR_CAN}             &2019 &1/64  \\
		BL \cite{Ma_2019_ICCV_BL}                    & 2019    &1/64  \\ 
		PACNN \cite{Shi_2019_CVPR_PACNN}       & 2019 &1/64  \\ 
		MCNN \cite{Zhang_2016_CVPR_MCNN}       & 2016 &1/16   \\ 
		Ours                   & - &1/4  \\ 
		\bottomrule
	\end{tabular} 
	\label{tab:resolution}
\end{table}

However, it is challenging for a vanilla deep neural network to produce a high-resolution density map. 
To make the network prediction robust, we introduce a multi-task loss by employing two separate upsampling branches for low-resolution and high-resolution supervision. As illustrated in Fig.~\ref{fig:network}, the losses of these two branches are supervised by the ground-truth density maps of two different scales, respectively. In particular, the low-resolution branch consists of two convolutional layers.
Hence, the network $\mathcal{F}$ is trained via a standard $L_2$ loss:
\begin{align}
	\mathcal{L}= \frac{1}{N} \sum_{i=1}^N &(\|\mathcal{F}_{HR}(X_i)-G_{HR}\|^2_2\nonumber\\ 
	&+ \|\mathcal{F}_{LR}(X_i)-G_{LR}\|^2_2),\label{eq:l2loss}
\end{align}
where $X_i$ denotes the $i^{th}$ input crowd image in the $N$ training images. $\mathcal{F}_{HR}$ and $\mathcal{F}_{LR}$ are the high-resolution and low-resolution outputs of our network. $G_{HR}$ and $G_{LR}$ represent the high-resolution and low-resolution ground-truth density maps, respectively. 

\subsection{Parametric perspective-aware density map}

The supervision of our crowd counting network requires the ground-truth crowd density map. 
Existing crowd counting benchmarks provide the point annotation for each crowd image, in which each point annotation represents the position of a person in the crowd. To obtain the crowd density map, following \cite{NIPS2010_4043_map}, prior crowd counting works convert the point annotation of a crowd image into a density map by applying a Gaussian kernel over each point. 
Most prior methods apply the Gaussian kernel with a fixed kernel size \cite{NIPS2010_4043_map,Cao_2018_ECCV_SA}, or the geometry-adaptive kernel size \cite{8099912_switch,Li_2018_CVPR_CSRNet,Liu_2019_CVPR_CAN,Liu_2019_ICCV_DSSINet,Ma_2019_ICCV_BL,xu2019learn,xiong2019open,shen2018crowd}. These density conversion techniques do not consider the perspective information or non-uniform crowd distribution of the crowd images. 
Based on the recent findings \cite{Wan_2019_ICCV_adaptive_map}, the generated density map may affect the performance of the model, so these trival density map generation methods may not help us achieve the satisfactory results. Thus, we propose a spatial-aware parametric method for generating adaptive density maps.

\begin{figure}
	\centering
	\setlength{\tabcolsep}{1pt}
	\begin{tabular}{cccc}
		\includegraphics[width=0.30\linewidth]{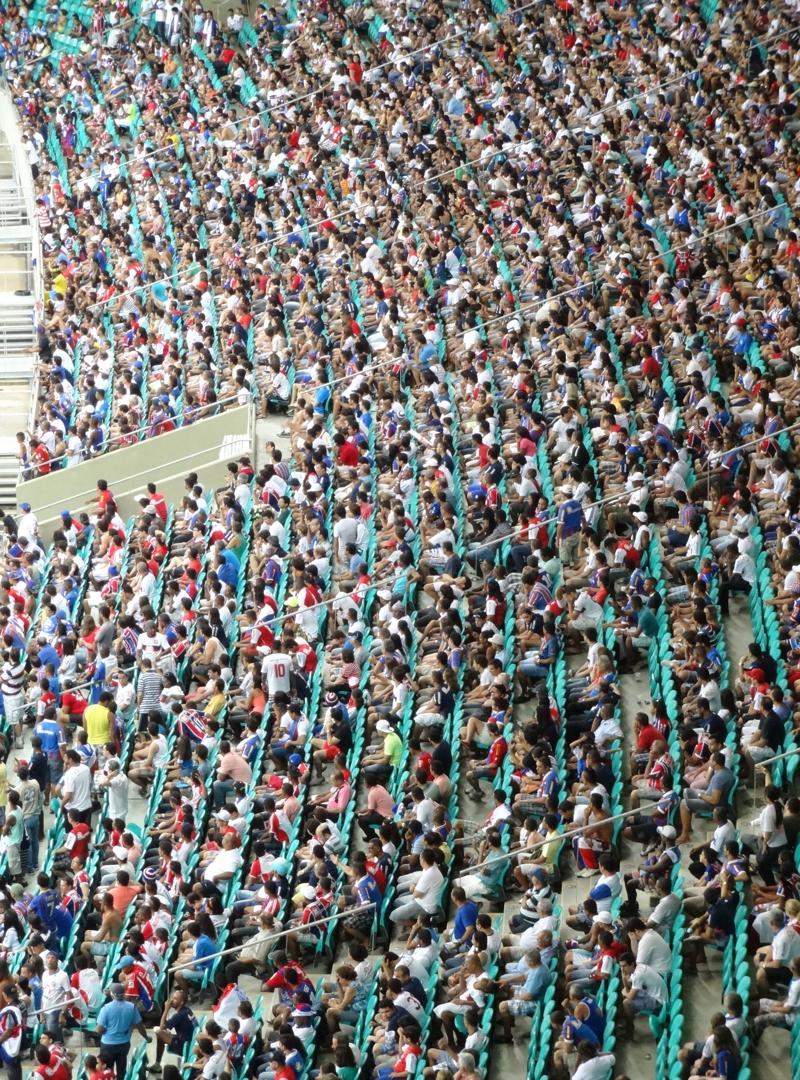}&
		\includegraphics[width=0.30\linewidth]{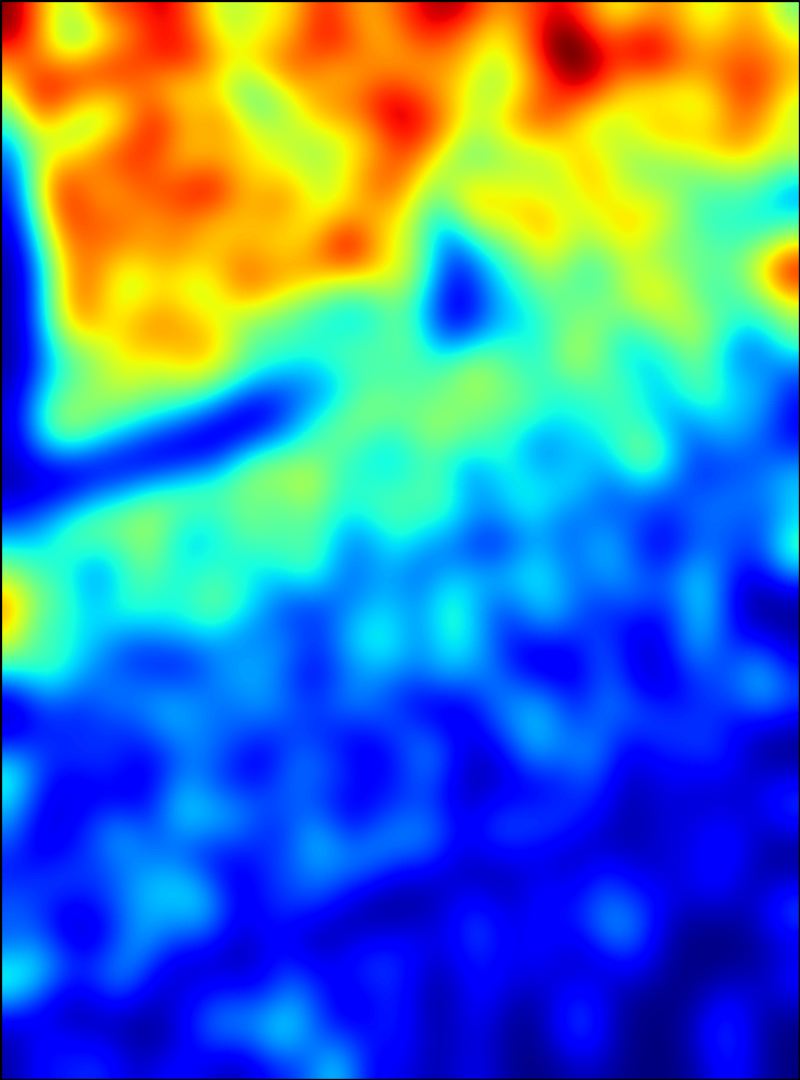}&
		\includegraphics[width=0.30\linewidth]{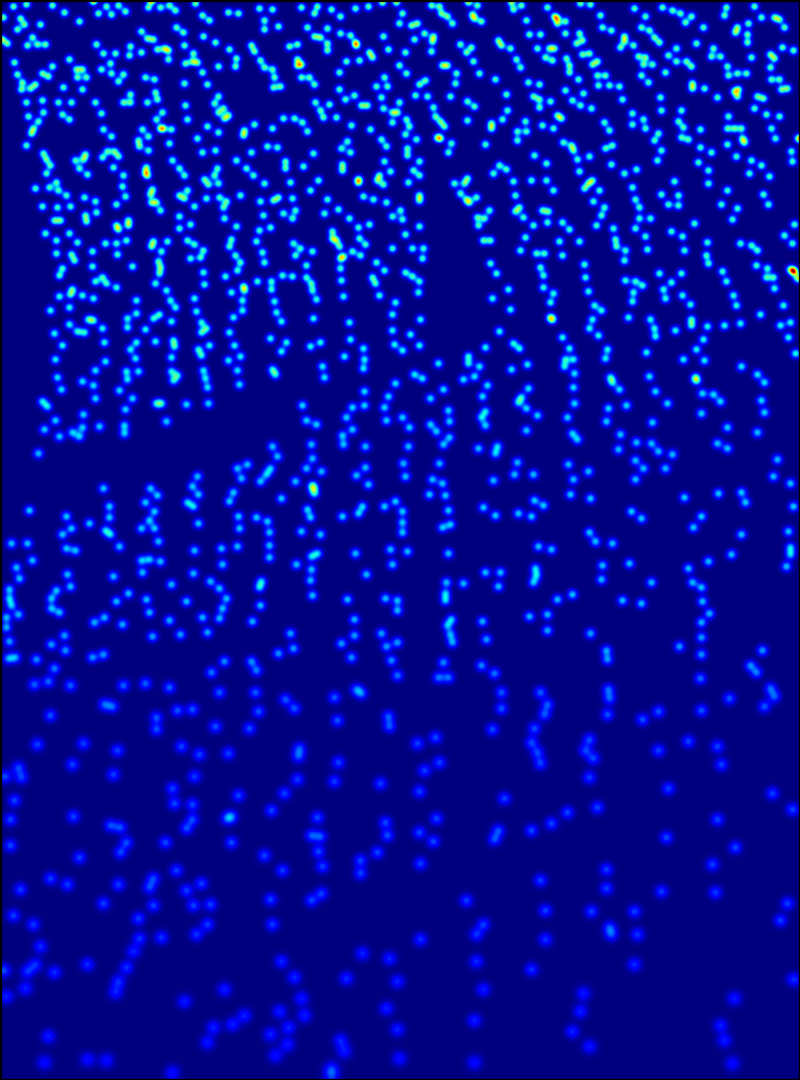}&	\includegraphics[trim=2.71cm 0 2.71cm 0, clip, width=0.06\linewidth]{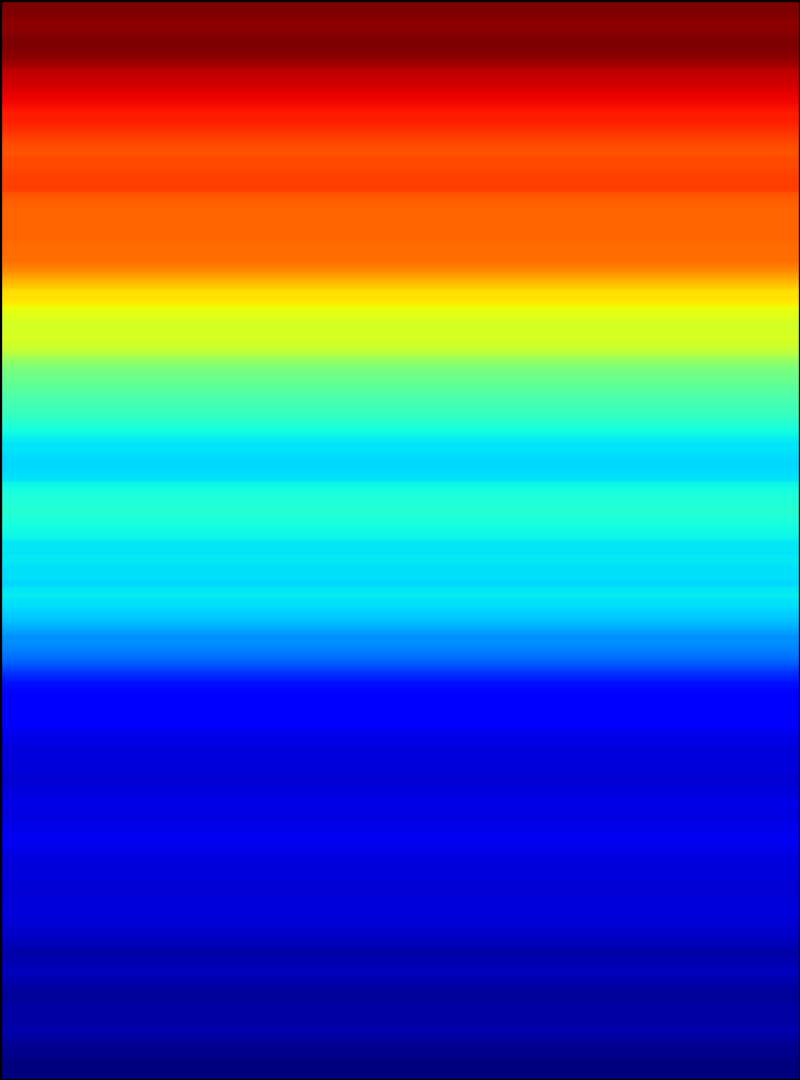}\\
		(a) & (b) & (c) &
	\end{tabular}
	\caption{(a) The input crowd image; (b) The prior density map $\bar{G}$ generated by a fixed and large Gaussian kernel size; (c) The density map generated by our proposed method. The rightmost color bar indicates the kernel size for each row.}
	\label{fig:density_map}
\end{figure}

Our proposed density conversion approach is based a simple assumption that the majority of persons located on the same row of the image have a similar distance away from the camera (e.g. Fig.~\ref{fig:density_map}(a)). 
This assumption may not be suited for certain crowd scenes, but it empirically works well for most scenes. Thus, subject to the assumption, each row of the density map should be corresponded to a Gaussian kernel with the same size, which can approximate the scale variance caused by the perspective view of the crowd image. 

To estimate the Gaussian kernel size for each row, we first apply a fixed large Gaussian kernel size (e.g., the standard deviation of Gaussian kernel $\sigma=25$) to produce a rough yet smooth density map $\bar{G}$ as a prior (see Fig.~\ref{fig:density_map}(b)). Due to the perspective effect of the image, the density concerns with not only the spatial distribution but also the distance away from camera. For instance, given two persons sitting side-by-side, they appear to be closer when they are farther away from camera and vice versa. Thus, we can approximately find out the relation between the density and the distance from camera for each row.
We propose to calculate the \textit{effective density} for the $i$-th row of $\bar{G}$:
\begin{align}
	D(i) &= \frac{1}{M}\sum_j \delta[\bar{G}(i,j) > \epsilon] \cdot \bar{G}(i,j),\\
	&\text{s.t. } M=\sum_j \delta[\bar{G}(i,j) > \epsilon],
\end{align}
where $\bar{G}(i,j)$ refers to the density value at the location $(i,j)$ of the image. $\delta[\cdot]$ denotes Dirac delta function, which equals to $1$ when the condition in the bracket is satisfied and $0$ otherwise. $\epsilon$ refers to the manually defined threshold. 

In essence, the \textit{effective density} measures the average density of most dense segment on each row of the crowd image while filtering out the less dense part. Thus, we can determine the kernel size for each row according to its $D(\cdot)$. To accomplish this, we apply a linear mapping between the maximum and minimum values of $D(\cdot)$ (i.e. $D_{max}$ and $D_{min}$) and 
the largest and smallest kernel size (i.e., the standard deviations $\sigma_{max}$ and $\sigma_{min}$), respectively. Specifically, $\sigma_{max}$ and $\sigma_{min}$ are manually determined, while $D_{max}$ and $D_{min}$ are measured over the entire training set. 
Hence, the Gaussian kernel size of the $i^{th}$ row can be computed as:
\begin{align}
	\sigma(i) = \alpha D^{-1}(i) + \beta,
\end{align}
where 
\begin{align}
	\beta &= \frac{D^{-1}_{min}\sigma_{min}-D^{-1}_{max}\sigma_{max}}{D^{-1}_{min}-D^{-1}_{max}},\\
	\alpha&= \frac{\sigma_{max}-\sigma_{min}}{D^{-1}_{min} - D^{-1}_{max}}.
\end{align}

Thus, after assigning the kernel size for each row, we can obtain the density map determined by $\sigma_{max}$ and $\sigma_{min}$, denoted as $G(\sigma_{min},\sigma_{max})$. We illustrate an example of our parametric method in Fig.~\ref{fig:density_map}(c). 
In practice, $\sigma_{min}$ is often fixed at a small value (e.g. $2.5$), while we mainly tune the value of $\sigma_{max}$ to achieve various density maps. Therefore, for the sake of simplicity, the notation of $G(\sigma_{min},\sigma_{max})$ can be simplified as $G(\sigma_{max})$. 
In the extreme case where $\sigma_{max}$ equals to $\sigma_{min}$, the generated density map is the same as the one generated by a universally-fixed kernel size. 

As described in the following section, our proposed density map generation technique can benefit the training of crowd counting model by providing a perspective-aware and parametric density map.

\subsection{Distillation based crowd counting}
\label{Sec:distillation}


As shown in Fig.~\ref{fig:density_map}(b) and (c), with a larger kernel size (i.e. larger $\sigma_{max}$ for $G(\sigma_{max})$), the density map will appear to be smoother, and vice versa. 
On one hand, for a deep neural network that regresses the density maps, the smoother density maps are easier to learn than the sharper ones. 
On the other hand, training the regression network from such smoother density maps may degrade the results of crowd counting, since the smoother density maps blur the individual information of the crowd image. In practice, proper parameters are often empirically chosen to trade off the counting accuracy and model training. 

\subsubsection{Crowd counting network distillation}

In order to break the bond caused by density maps, we propose a distillation based optimization method that enables to progressively reduce the difficulty of learning from sharper density maps so as to obtain a better solution. Knowledge distillation has been proposed in \cite{hinton2015distilling}, which has been applied for network compression. Specificially, a lightweight model is often applied to learn the behavior of a large network. The recent studies find that the identical network structures can benefit from the distillation \cite{furlanello2018born,yang2019snapshot}. Here, we employ the similar optimization strategy to iteratively train our network.

First, we train our network according to Eq.~\ref{eq:l2loss} that can be simply expressed as:
\begin{align}
	\mathcal{L}^{(t=0)}=  \frac{1}{N}{\sum_{i=1}^N\|\mathcal{F}(X_i)-G\|^2}.\label{eq:init}
\end{align}
Next, we treat the trained network $\mathcal{F}$ as the teacher model (denoted as $\mathcal{F}_T$) and we leverage it to train an identical network (i.e., the student model $\mathcal{F}_S$) from scratch. Thus, the trained student model at this stage can be treated as the teacher model of the next stage, i.e. $\mathcal{F}^{(t)}_T \leftarrow \mathcal{F}^{(t-1)}_S$, where $t$ denotes the timestamp of the training stage. Hence, the training loss of the $t^{th}$ stage in distillation can be expressed as follows:
\begin{align}
	\mathcal{L}^{(t)}= & \frac{1}{N}{\sum_{i=1}^N\|\mathcal{F}^{(t)}_S(X_i)-G\|^2} + \nonumber\\ 
	&  \frac{\lambda}{N} {\sum_{i=1}^N\|\mathcal{F}^{(t)}_S(X_i)-\mathcal{F}^{(t)}_T(X_i)\|^2},\label{eq:distill}
\end{align}
where $\lambda$ denotes the constant balance weight. The first term is a standard $L2$ loss, while the second term aims to align the outputs of the teacher model and the student model, which forces the student model to approach the behavior of the teacher model.

\begin{figure}
	\centering
	\includegraphics[width=0.5\textwidth]{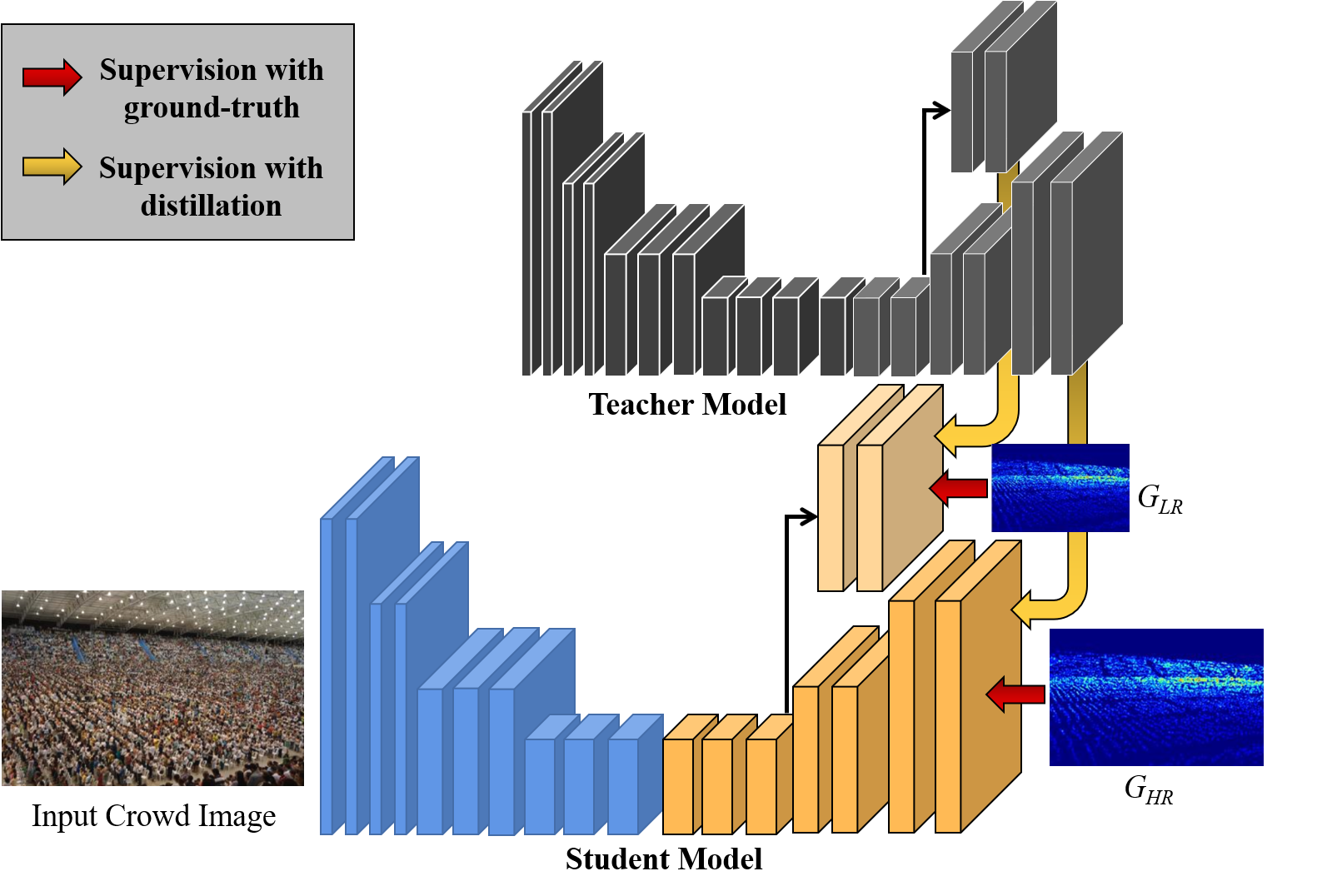}
	\caption{Illustration of the distillation-based optimization for our crowd counting model. As shown, we first train a teacher model and freeze its weights. Then, we utilize the teacher model to train another model (i.e., the student model) with identical structure supervised by the low-resolution and high-resolution ground-truth density maps.}
	\label{fig:distill}
\end{figure}

The distillation based optimization can progressively optimize the network. However, as we mentioned, the inappropriately computed ground-truth density maps will degrade the counting performance. Furthermore, the static ground-truth density maps (i.e., $G$ remains the same in distillation) limit the optimal performance discovered by the distillation. Thus, in the following section, we introduce a simple yet effective method in order to produce adaptive density maps.

\subsubsection{Density-aware distillation}

To further strengthen the network performance, utilizing our parameteric perspective-aware density map generation technique, we improve the distillation method. According to Eq.~\ref{eq:distill}, the student model $\mathcal{F}_S$ will learn the behavior of the teacher model $\mathcal{F}_T$. But, with the static density map $G$ as ground-truth, the distillation may not further improve the performance. However, it is difficult to directly train a network from sharp density maps generated with a small kernel size, which often leads to a poor performance. With distillation, the model is able to progressively adapt to the sharper and sharper density map.
Particularly, in each stage of distillation, we slightly increase the difficulty of the task by introducing the density maps with a smaller kernel size. 
In specific, on every stage of distillation, the training loss is modified as below:
\begin{align}
	\mathcal{L}^{(t)}= & \frac{1}{N}{\sum_{i=1}^N\|\mathcal{F}^{(t)}_S(X_i)-G^{(t)}(\sigma^{(t)}_{max})\|^2} \nonumber\\ 
	&  +\frac{\lambda}{N} {\sum_{i=1}^N\|\mathcal{F}^{(t)}_S(X_i)-\mathcal{F}^{(t-1)}_T(X_i)\|^2},\label{eq:tsoptim}
\end{align}
where $\sigma^{(t)}_{max}$ refers to the maximal Gaussian kernel size at the $t^{th}$ training stage. After each stage, $\sigma^{(t)}_{max}$ will be updated, i.e., $\sigma^{(t)}_{max}=w \sigma^{(t-1)}_{max}$, where $w$ is a constant within the value range of $[0,1)$. Our algorithm is summarized in Algorithm~\ref{algo}. By iteratively distilling the model while reducing the kernel size to generate sharper density maps for network to learn, the performance of the student model can iteratively be strengthened. 

\begin{algorithm}
	\SetAlgoLined
	\KwIn{$\text{Network }\mathcal{F}, \text{Input images }X_i (i=1\cdots N)$, $\text{Maximal kernel size }\sigma^{(0)}_{max}$}
	Compute density map $G$ using $\sigma^{(0)}_{max}$ (i.e. $G^{(t=0)}$)\;
	Train the initial model $\mathcal{F}$ supervised by $G$ (Eq.~\ref{eq:init})\;
	Assign the teacher model $\mathcal{F}_T^{(t=0)} \leftarrow \mathcal{F}$\;
	\For{t=1:T}{
		$\sigma^{(t)}_{max} \leftarrow w\sigma^{(t-1)}_{max}$\;
		Produce density map $G^{(t)}$ using $\sigma^{(t)}_{max}$\;
		Set up a new student model $\mathcal{F}_S^{(t)}$\;
		Train $\mathcal{F}_S^{(t)}$ using $G^{(t)}$ and $\mathcal{F}_T^{(t-1)}$ (Eq.~\ref{eq:tsoptim})\;	
		$\mathcal{F}_T^{(t)} \leftarrow \mathcal{F}_S^{(t)}$\;
	}
	\caption{Iterative distillation of crowd counting}
	\label{algo}
\end{algorithm}

\section{Experimental Results}
\label{sec:exp}

In this section, we first introduce the datasets used for evaluation and the metrics, as well as the implementation details. Then, we conduct the comparison experiments with the state-of-the-art methods in public benchmarks. Last, we perform the ablation study to investigate the density map generation, the distillation algorithm, and the multi-task loss of our model.

\subsection{Datasets and evaluation metrics}

We evaluate our approach on public crowd counting benchmarks: ShanghaiTech Part A/B~\cite{Zhang_2016_CVPR_MCNN}, UCF-QNRF~\cite{Idrees_2018_ECCV_CL}, and UCSD~\cite{chan2008privacy}.
Particularly, ShanghaiTech dataset has a total of 1198 crowd images, including Part A and Part B. Part A contains 482 images, and Part B contains 716 images. The QNRF dataset has 1535 images with average resolution at $2013\times 2902$ and a total of 1.25 million annotations, in which there are 1201 images for training and 334 images for testing. Besides, the UCSD dataset have 2000 images with the resolution of $238\times 158$ with relatively smaller density.


Following prior works, MSE and MAE are used as metrics, which are defined as follows:
\begin{align}
	MAE &= \frac{1}{N} \sum_{i=1}^N\|E_i-GT_i\|_1,\\
	MSE &= \frac{1}{N} \sqrt{\sum_{i=1}^N\|E_i-GT_i\|_2^2},
\end{align}
where $N$ is the number of test images, $E_i$ indicates the estimated density map count of the $i$-th image and $GT_i$ indicates the ground-truth count of the $i$-th image.

\subsection{Implementation details}

In practice, we set $\sigma_{min}$ as 2.5 and the initial $\sigma_{max}$ as 25. In the first timestep, the $\sigma_{max}$ is set as 20 and $w$ as 0.5. 
On the initialization stage, the learning rate is set as $5\times 10^{-6}$ and the momentum 0.95. Since we do not normalize the input image, the batch size is set as 1. Other training settings differ across the benchmarks, so we will elaborate them below, respectively. 

\begin{table}
	\setlength{\tabcolsep}{4pt}
	\centering
	\caption{The comparison results on the ShanghaiTech Part A (SHA) and Part B (SHB). }
	\begin{tabular}{c|c|c|c|c|c}
		\toprule
		\multirow{2}{*}{Method}  & \multirow{2}{*}{Year}& \multicolumn{2}{c|}{SHA}& \multicolumn{2}{c}{SHB} \\
		\cmidrule{3-6}
		& &  MAE& MSE & MAE & MSE \\
		\midrule
		\midrule
		MCNN~\cite{Zhang_2016_CVPR_MCNN}  & 2016 & 110.2 & 173.2 & 26.4 & 41.3\\
		Switching CNN~\cite{8099912_switch} &2017 & 90.4 & 135.0 & 21.6& 33.4 \\
		SANet~\cite{Cao_2018_ECCV_SA} & 2018& 67.0 & 104.5 & 8.4& 13.6\\
		CSRNet~\cite{Li_2018_CVPR_CSRNet}    &2018& 68.2 & 115.0 & 10.6 & 16.0 \\
		ic-CNN~\cite{ranjan2018iterative} & 2018 & 69.8& 117.3& 10.4 & 16.7\\
		CSRNet+PACNN~\cite{Shi_2019_CVPR_PACNN}   &2019             &62.4  &102.0 & 7.6 & 11.8\\ 
		ADCrowdNet~\cite{Liu_2019_CVPR_ADCrowd} &2019&63.2  &98.9 & 7.6 & 13.9 \\
		PACNN~\cite{Shi_2019_CVPR_PACNN} & 2019& 66.3 & 106.4 & 8.9& 13.5\\
		CAN~\cite{Liu_2019_CVPR_CAN}   & 2019             &62.3  & 100.0 &7.8 &12.2 \\
		BL~\cite{Ma_2019_ICCV_BL}            & 2019              &62.8  & 101.8 &7.7&12.7 \\
		TEDnet~\cite{jiang2019crowd} & 2019 & 64.2 & 109.1 & 8.2 & 12.8\\
		HA-CCN~\cite{sindagi2019ha} & 2019& 62.9 & \textbf{94.9} & 8.1 & 13.4\\
		Ours                       & - &\textbf{61.1} &104.7 &\textbf{7.5}&\textbf{12.0} \\ 
		\bottomrule 
		
	\end{tabular}

	\label{Tab:Shanghaitech}	
\end{table}

\begin{table}
	\setlength{\tabcolsep}{6pt}
	\centering
	\caption{Comparison of our approach with other state-of-the-art methods on UCF-QNRF.}
	\begin{tabular}{c|c|c|c}
		\toprule
		Method & Year & MAE& MSE\\
		\midrule
		\midrule
		MCNN~\cite{Zhang_2016_CVPR_MCNN} & 2016 & 277.0 & 426.0\\
		Switching CNN~\cite{8099912_switch} & 2017 & 228.0 & 445.0\\
		CL~\cite{Idrees_2018_ECCV_CL}           & 2018    &132.0   &191.0 \\
		HA-CCN~\cite{sindagi2019ha} & 2019 & 118.1 & 180.4\\
		RANet~\cite{zhang2019relational} & 2019 & 111.0 & 190.0\\
		CAN~\cite{Liu_2019_CVPR_CAN}             & 2019   &107.0   &183.0 \\
		TEDnet~\cite{jiang2019crowd} & 2019 & 113.0 & 188.0\\
		SPN+L2SM~\cite{xu2019learn} & 2019 & 104.7 & 173.6\\
		S-DCNet~\cite{xiong2019open} &	2019 & 104.4	&	176.1\\
		SFCN~\cite{wang2019learning}	& 2019 & 102.0	&	171.4\\
		DSSINet~\cite{Liu_2019_ICCV_DSSINet}     & 2019       &99.1  &\textbf{159.2}  \\  
		MBTTBF-SCFB~\cite{Sindagi_2019_ICCV_MBTTBF}    & 2019    &97.5  &165.2 \\ 
		Ours             & -           &\textbf{92.9} &\textbf{159.2}  \\ 
		\bottomrule
	\end{tabular}
	\label{Tab:UCF-QNRF}
\end{table}

\begin{table}
	\setlength{\tabcolsep}{6pt}
	\centering
	\caption{The comparison results on UCSD. }
	\begin{tabular}{c|c|c|c}
		\toprule
		Method & Year & MAE& MSE\\
		\midrule
		\midrule
		MCNN~\cite{Zhang_2016_CVPR_MCNN} & 2016 & 1.07 & 1.35\\
		Switching CNN~\cite{8099912_switch} & 2017 & 1.62 & 2.10\\
		ConvLSTM~\cite{xiong2017spatiotemporal} & 2017 & 1.30 & 1.79\\
		BSAD~\cite{huang2017body} & 2017 & 1.00 & 1.40\\
		ACSCP~\cite{shen2018crowd} & 2018 & 1.04 & 1.35\\
		CSRNet~\cite{Li_2018_CVPR_CSRNet}   & 2018 &1.16  &1.47  \\ 	 
		SANet~\cite{Cao_2018_ECCV_SA}             & 2018         &1.02  &1.29  \\  SANet+SPANet~\cite{Cheng_2019_ICCV_SPANet}  & 2019              &1.00  &1.28 \\  
		ADCrowdNet~\cite{Liu_2019_CVPR_ADCrowd}         & 2019    &\textbf{0.98}  &1.25 \\ 
		Ours         &     -          &\textbf{0.98}  &  \textbf{1.24}\\ 
		\bottomrule
	\end{tabular}
	
	\label{Tab:UCSD}
	\vspace{-0.4cm}
\end{table}

\begin{figure*}
	\centering
	\setlength{\tabcolsep}{1pt}
	\begin{tabular}{ccccc}
		Input image & Bayesian~\cite{Ma_2019_ICCV_BL} & CSRNet~\cite{Li_2018_CVPR_CSRNet} & Ours & Ground-truth\\
		\includegraphics[width=0.2\textwidth]{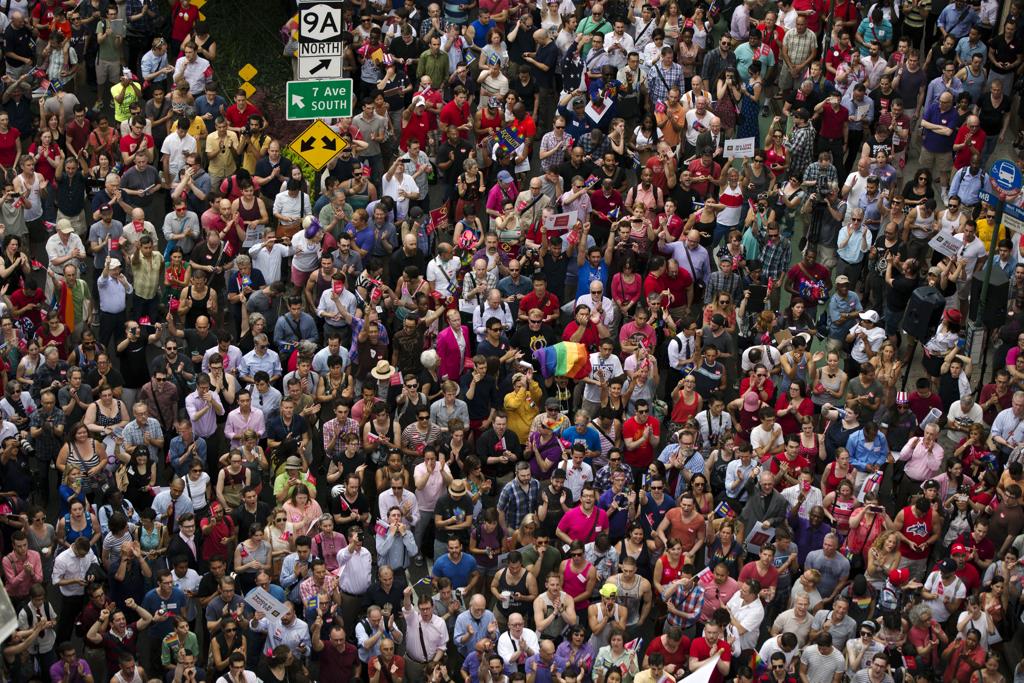}&
		\includegraphics[width=0.2\textwidth]{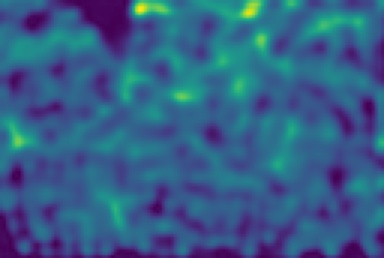}&
		\includegraphics[width=0.2\textwidth]{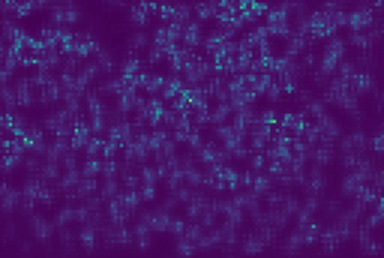}&
		\includegraphics[width=0.2\textwidth]{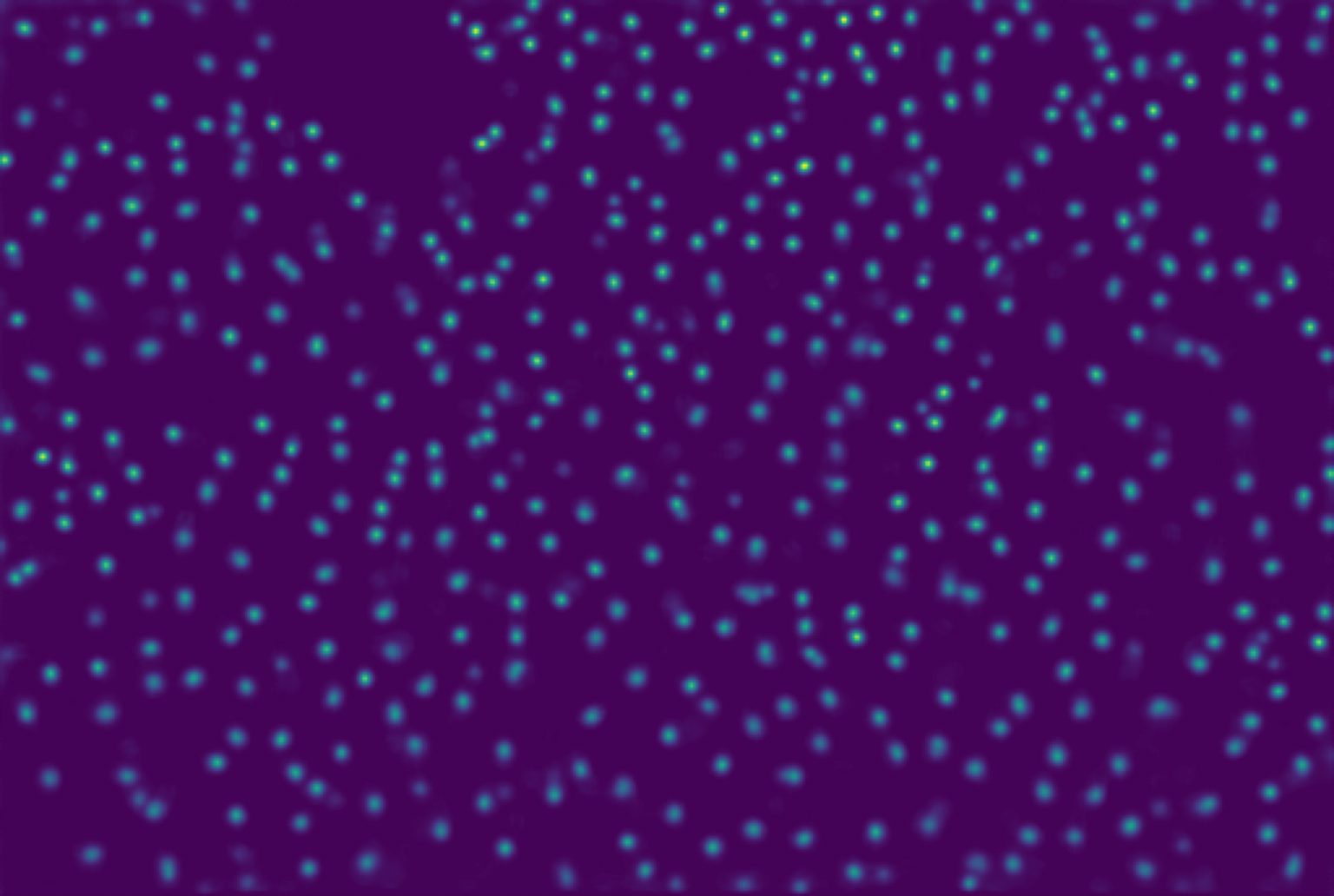}&
		\includegraphics[width=0.2\textwidth]{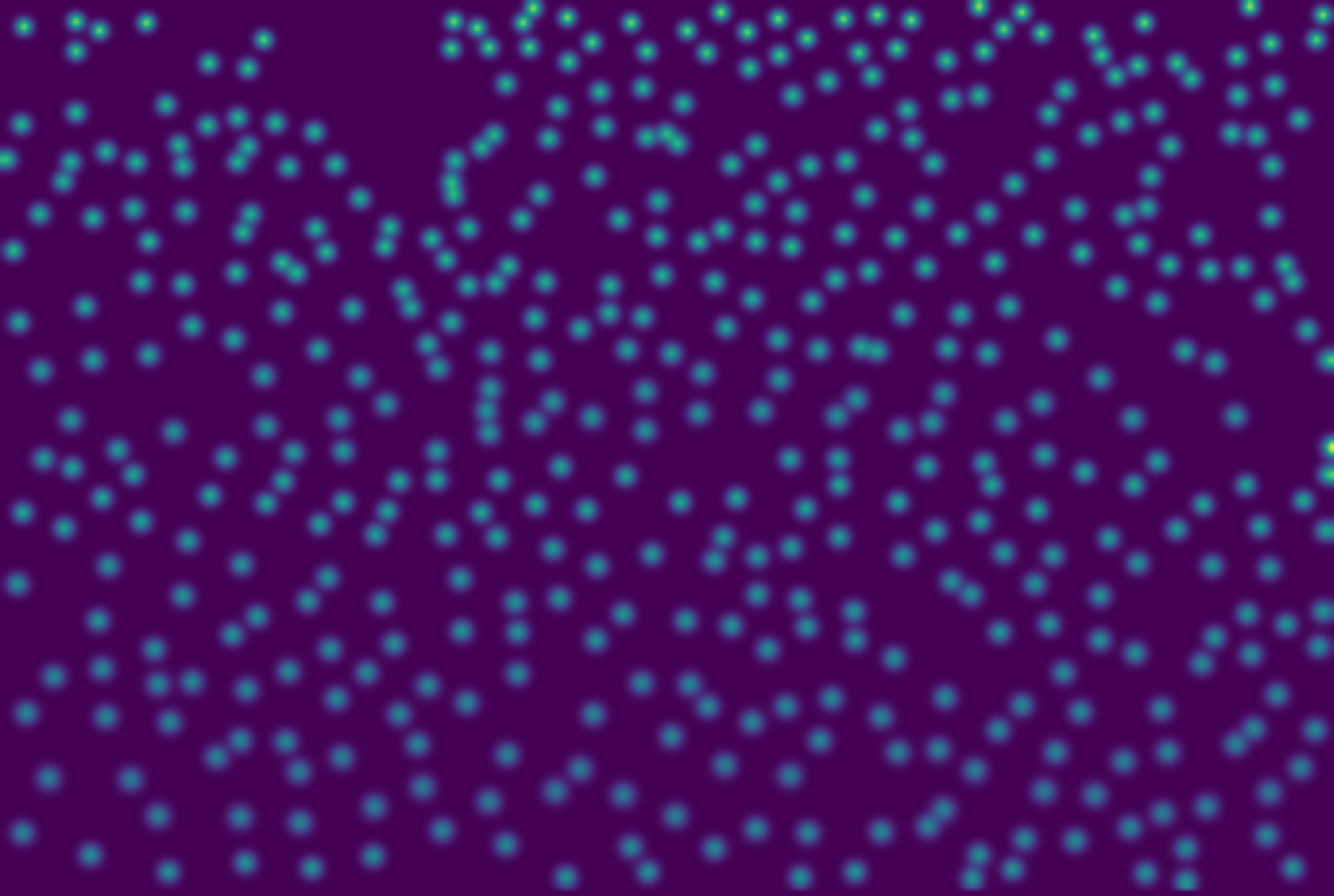}\\
		Estimated count & 464.4 & 481.5 & 460.4 & 460\\
		\includegraphics[width=0.2\textwidth]{images/dif_method/IMG_22.jpg}&
		\includegraphics[width=0.2\textwidth]{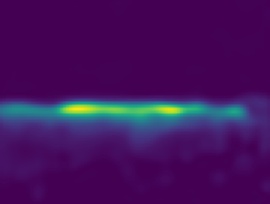}&
		\includegraphics[width=0.2\textwidth]{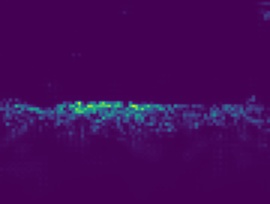}&
		\includegraphics[width=0.2\textwidth]{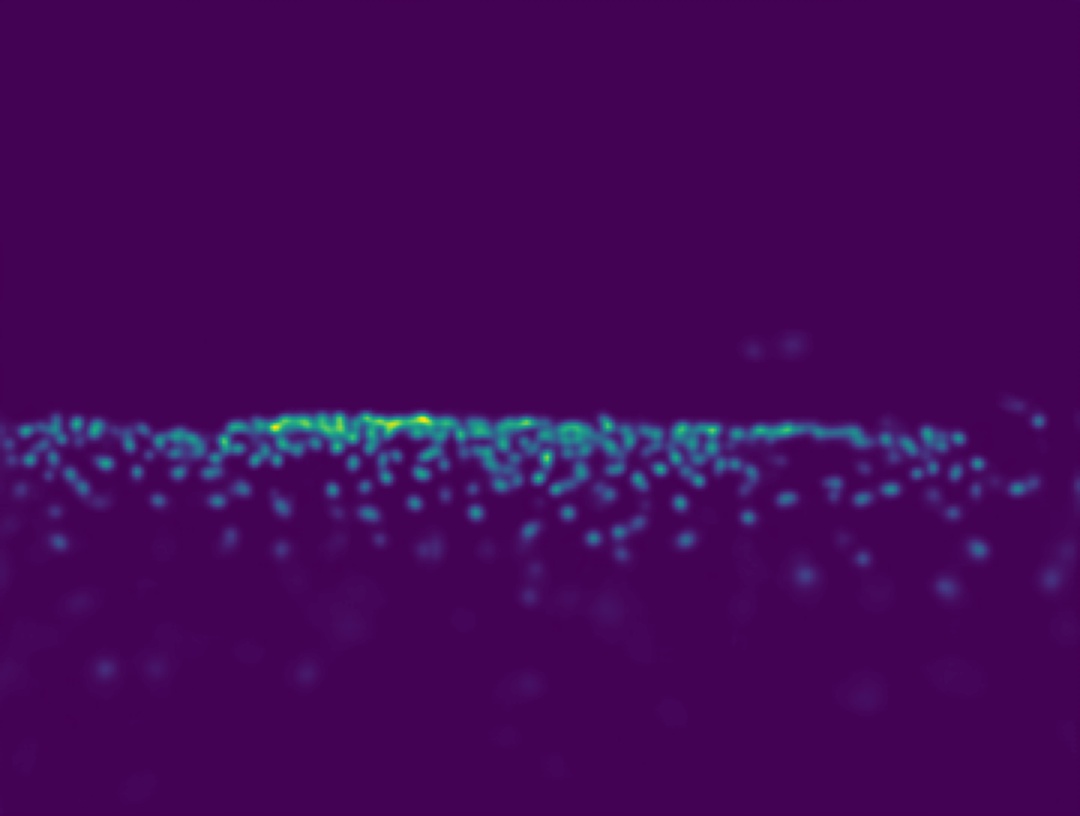}&
		\includegraphics[width=0.2\textwidth]{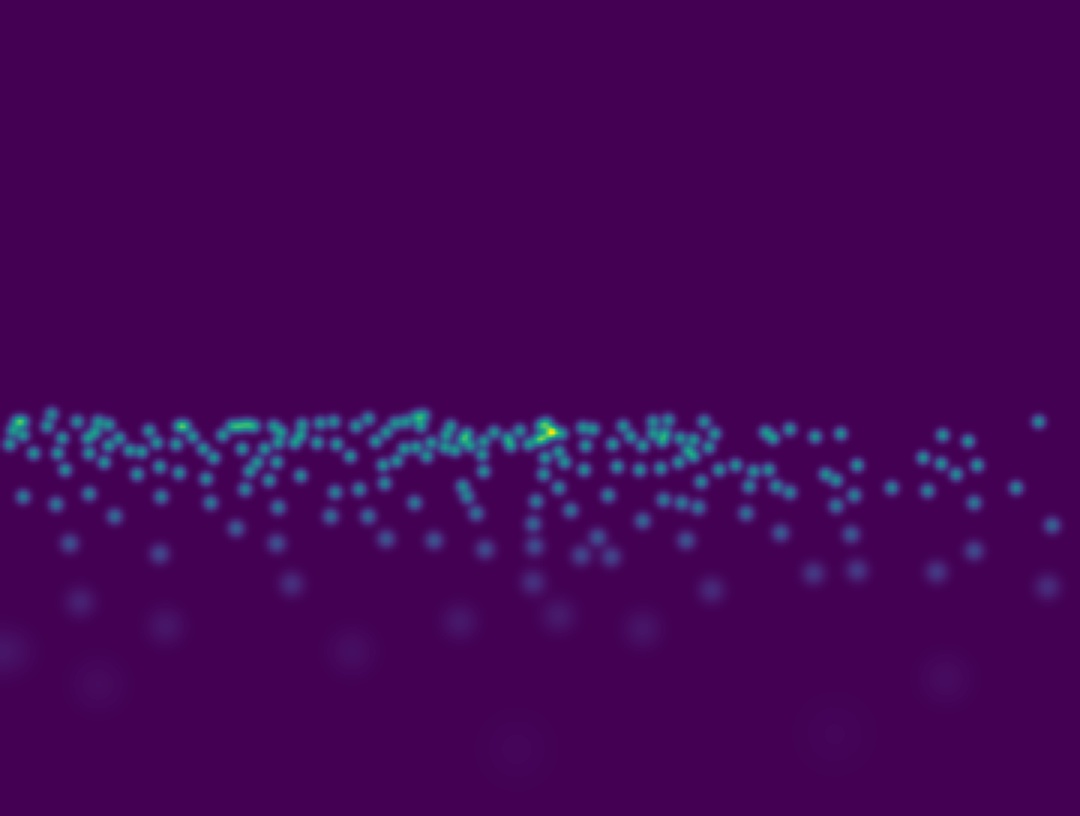}\\
		Estimated count & 269.9 & 274.4 & 231.7 & 212\\
		\includegraphics[width=0.2\textwidth]{images/dif_method/IMG_27.jpg}&
		\includegraphics[width=0.2\textwidth]{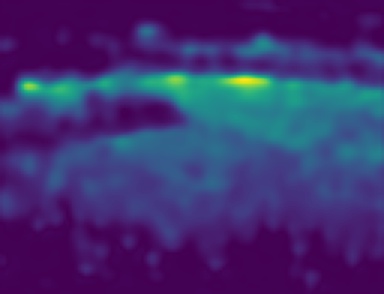}&
		\includegraphics[width=0.2\textwidth]{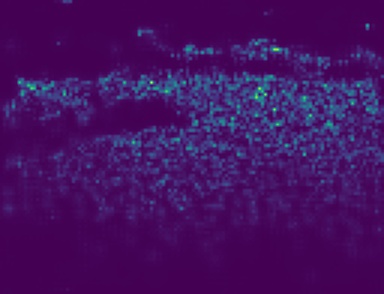}&
		\includegraphics[width=0.2\textwidth]{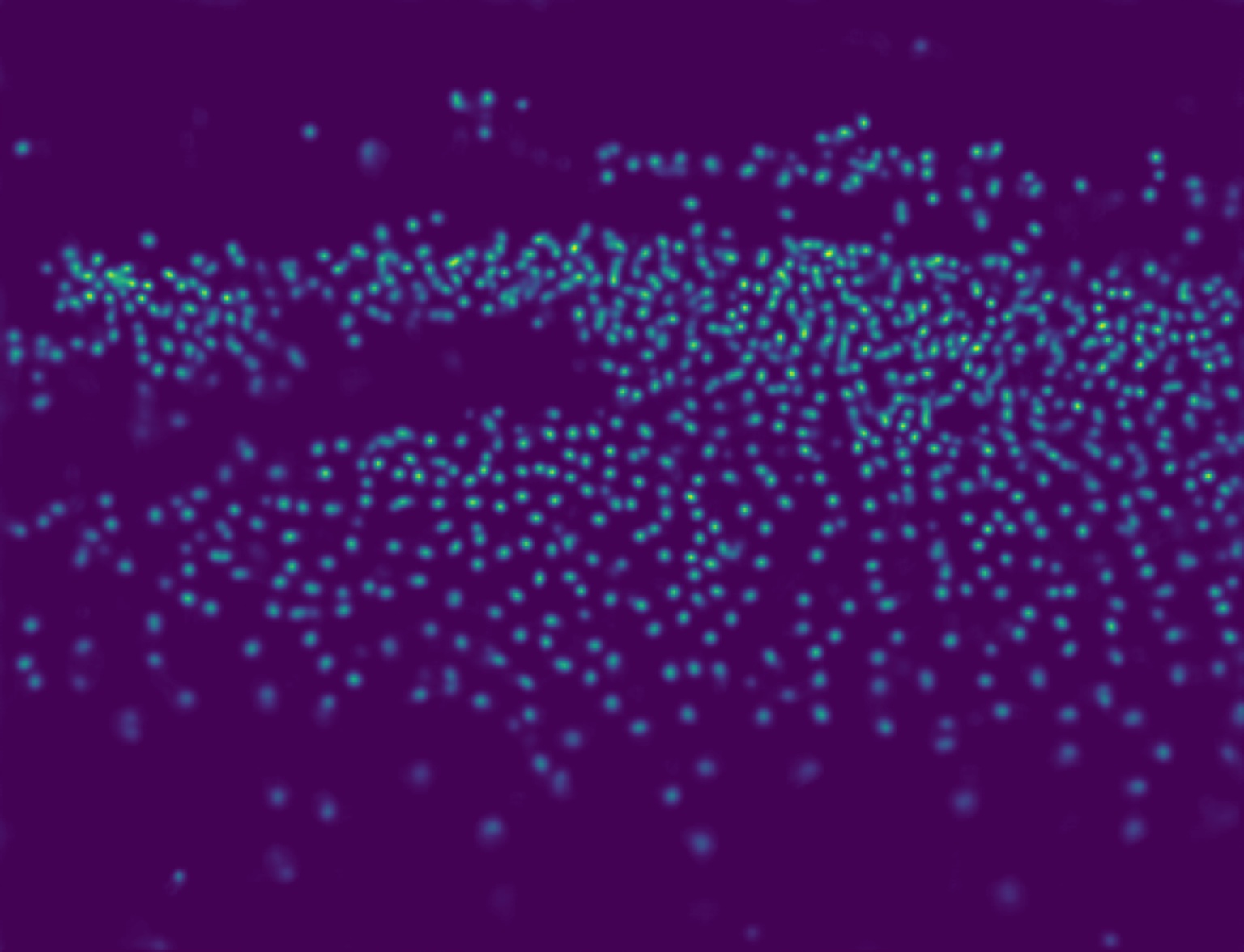}&
		\includegraphics[width=0.2\textwidth]{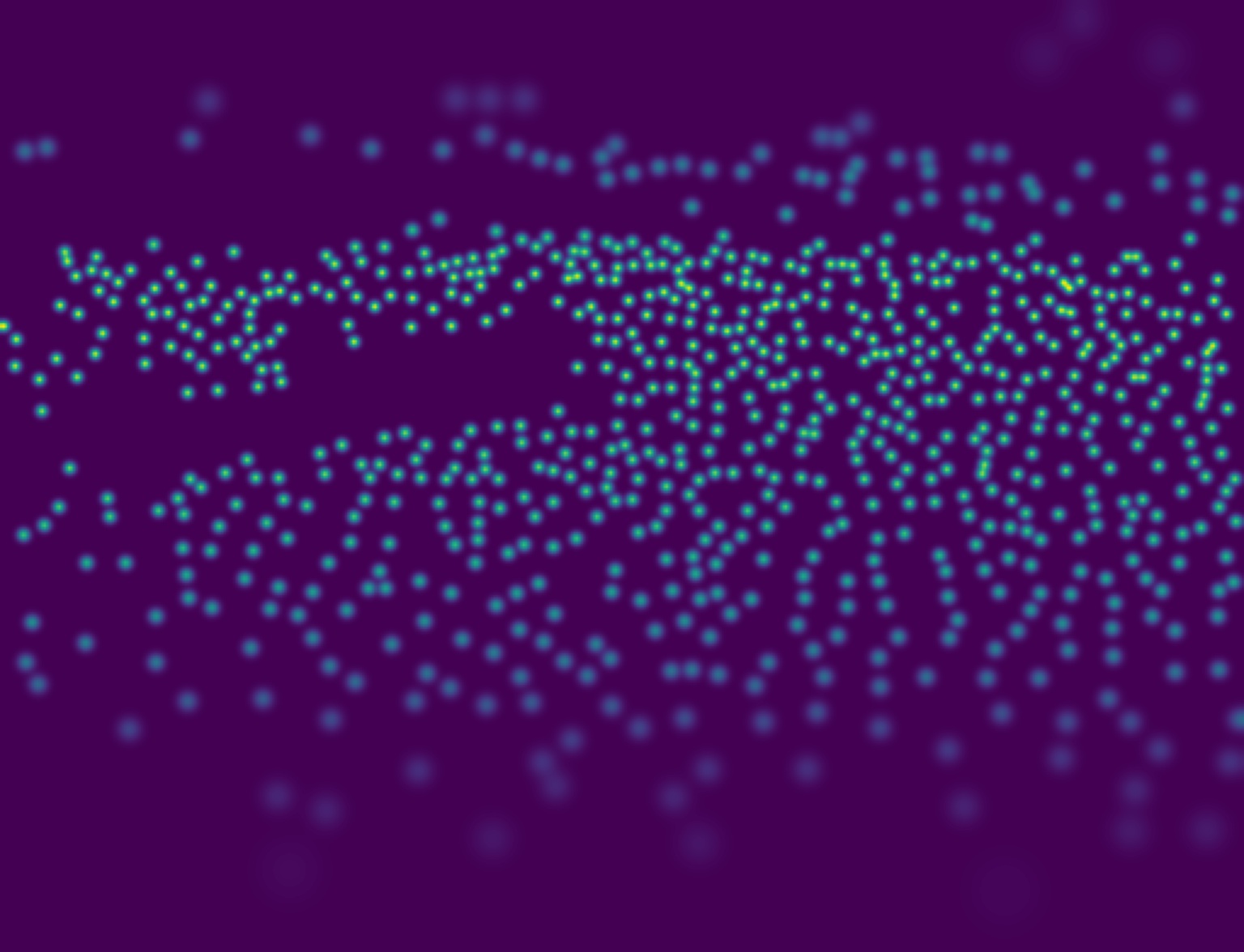}\\
		Estimated count & 797.8 & 917.5 & 758.1 & 760\\
		\includegraphics[width=0.2\textwidth]{images/dif_method/IMG_28.jpg}&
		\includegraphics[width=0.2\textwidth]{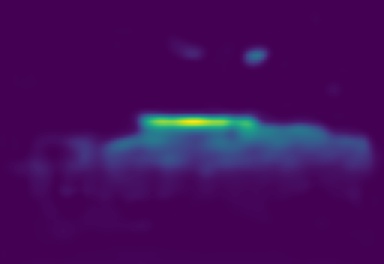}&
		\includegraphics[width=0.2\textwidth]{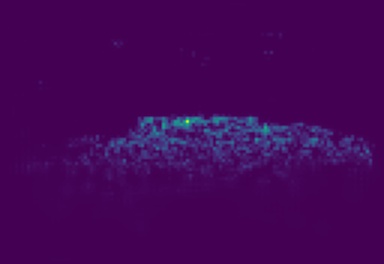}&
		\includegraphics[width=0.2\textwidth]{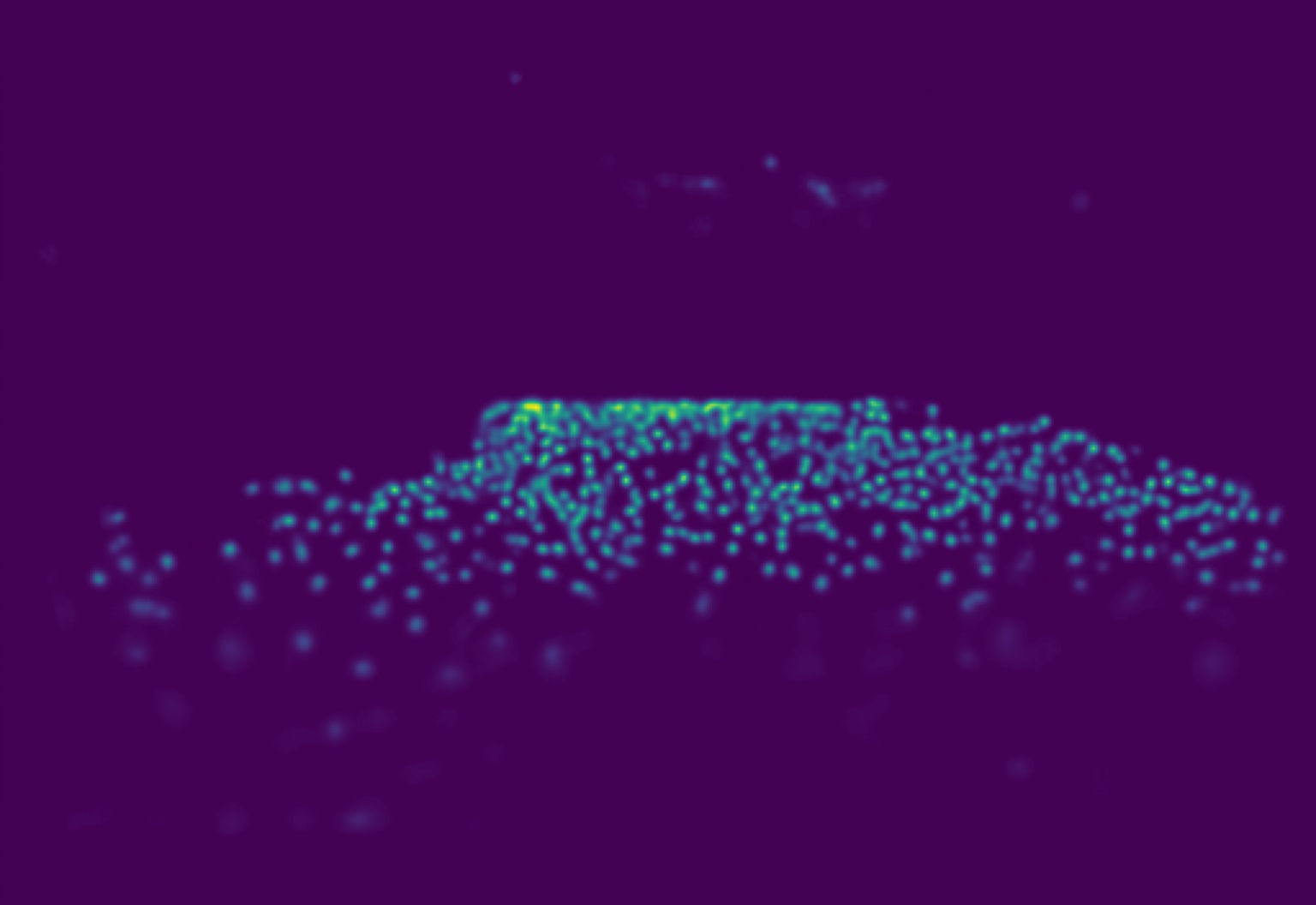}&
		\includegraphics[width=0.2\textwidth]{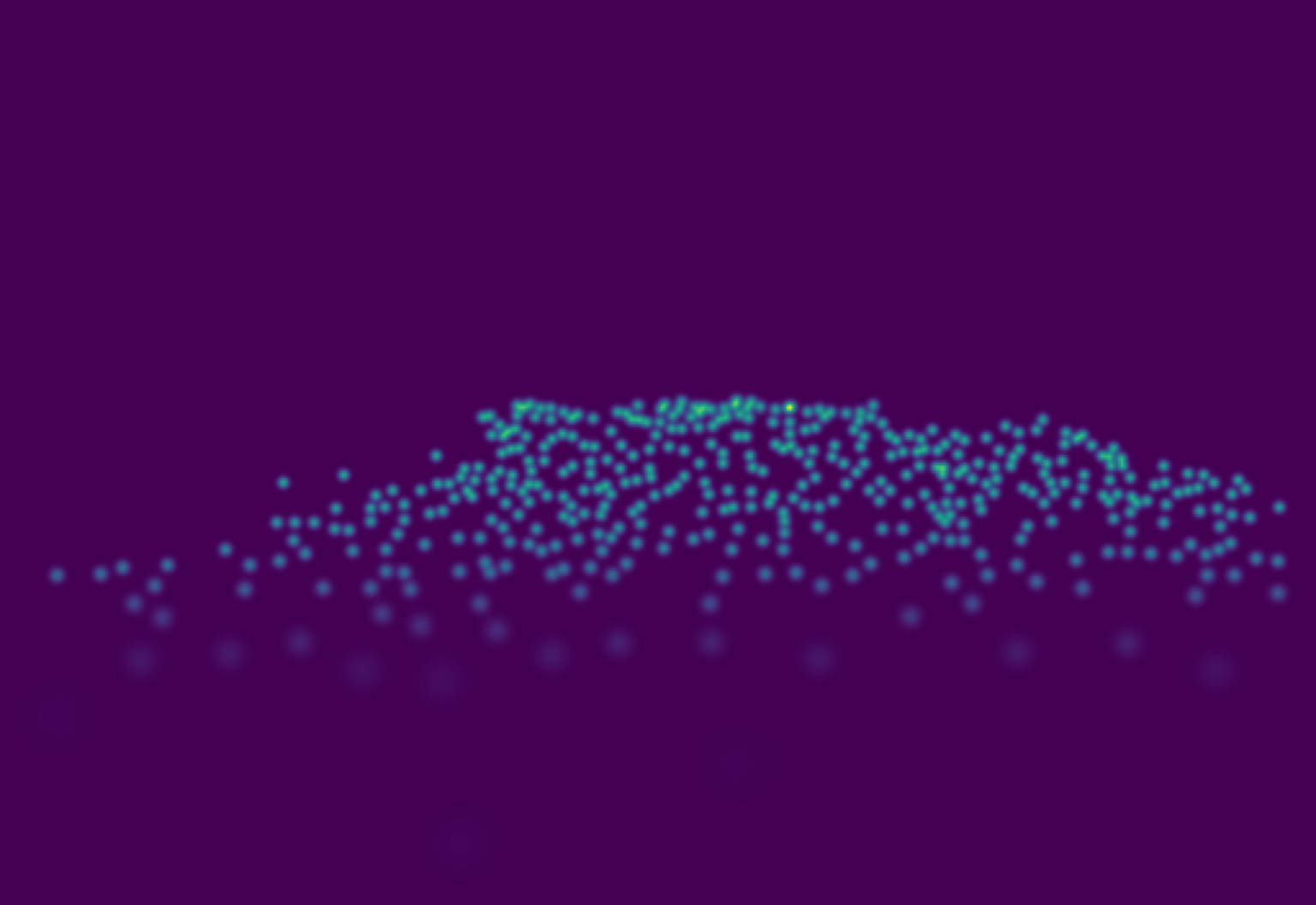}\\
		Estimated count & 542.7 & 556.3 & 440.2 & 423\\
		\includegraphics[width=0.2\textwidth]{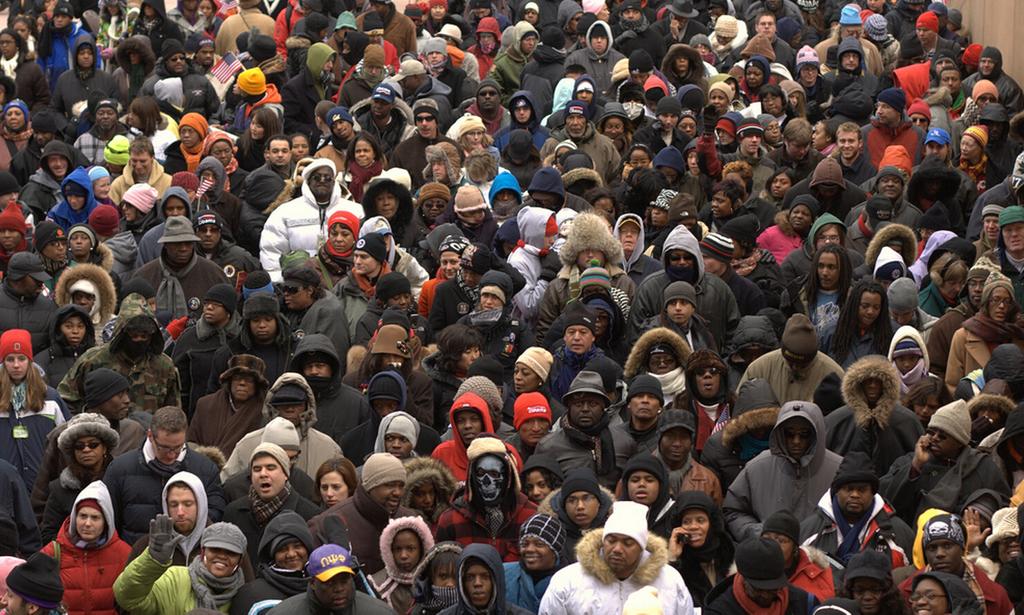}&
		\includegraphics[width=0.2\textwidth]{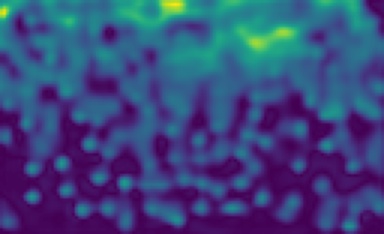}&
		\includegraphics[width=0.2\textwidth]{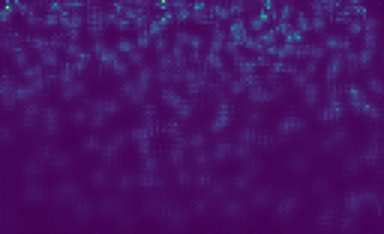}&
		\includegraphics[width=0.2\textwidth]{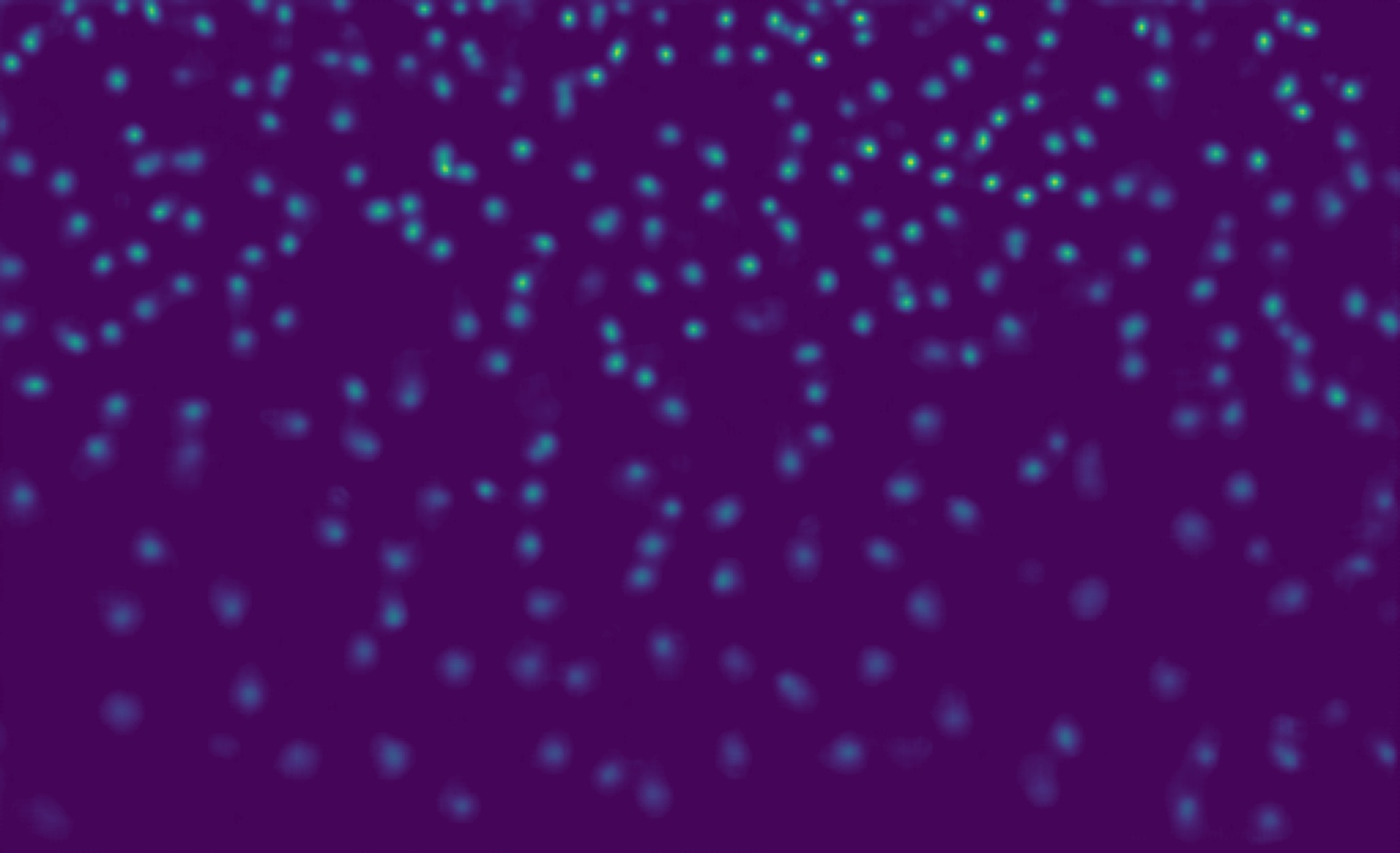}&
		\includegraphics[width=0.2\textwidth]{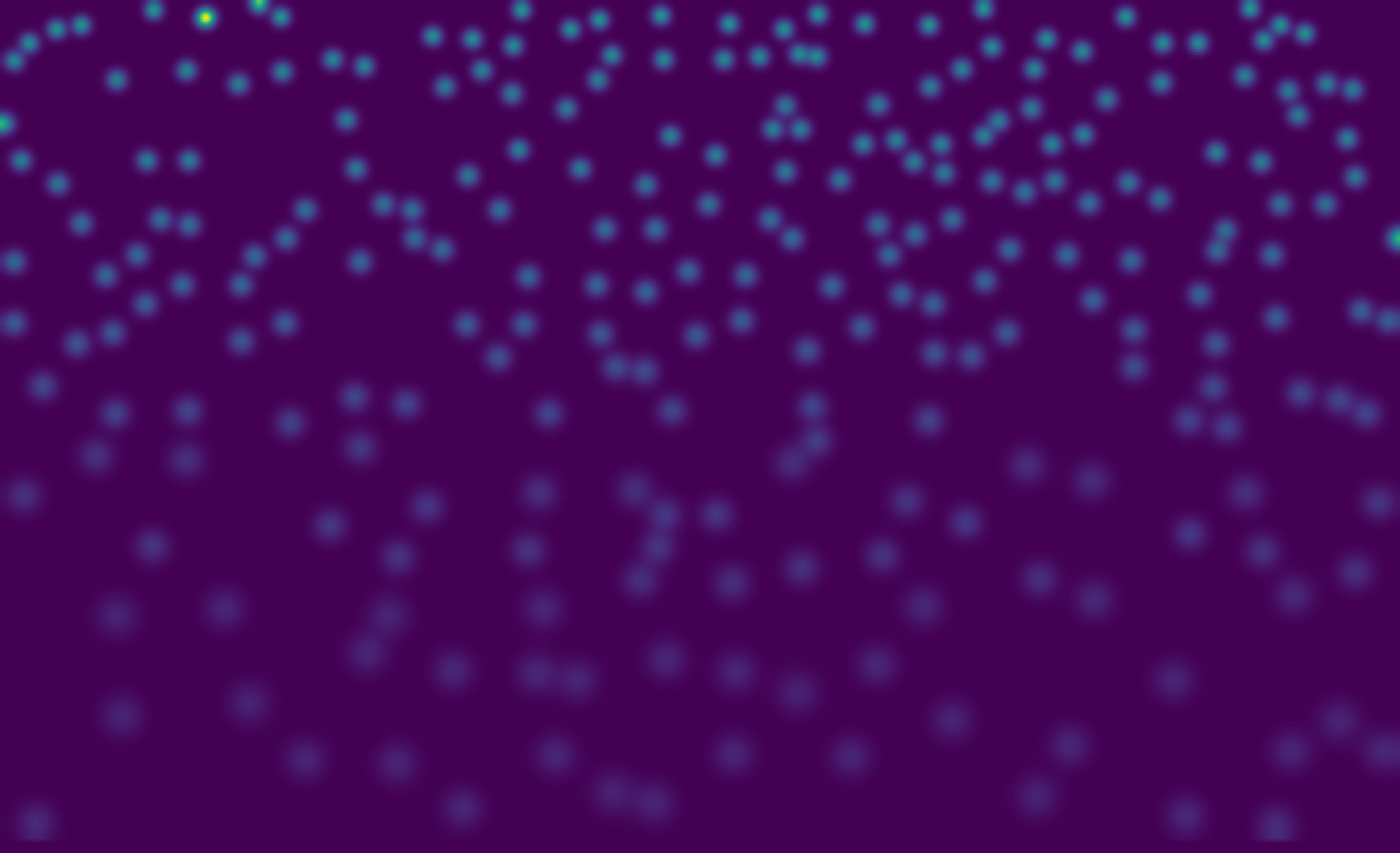}\\
		Estimated count & 260.5 & 255.7 & 252.2 & 250\\
		\includegraphics[width=0.2\textwidth]{images/dif_method/IMG_46.jpg}&
		\includegraphics[width=0.2\textwidth]{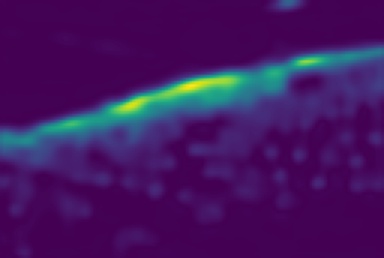}&
		\includegraphics[width=0.2\textwidth]{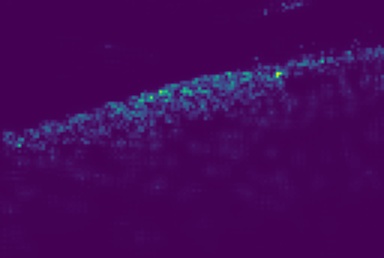}&
		\includegraphics[width=0.2\textwidth]{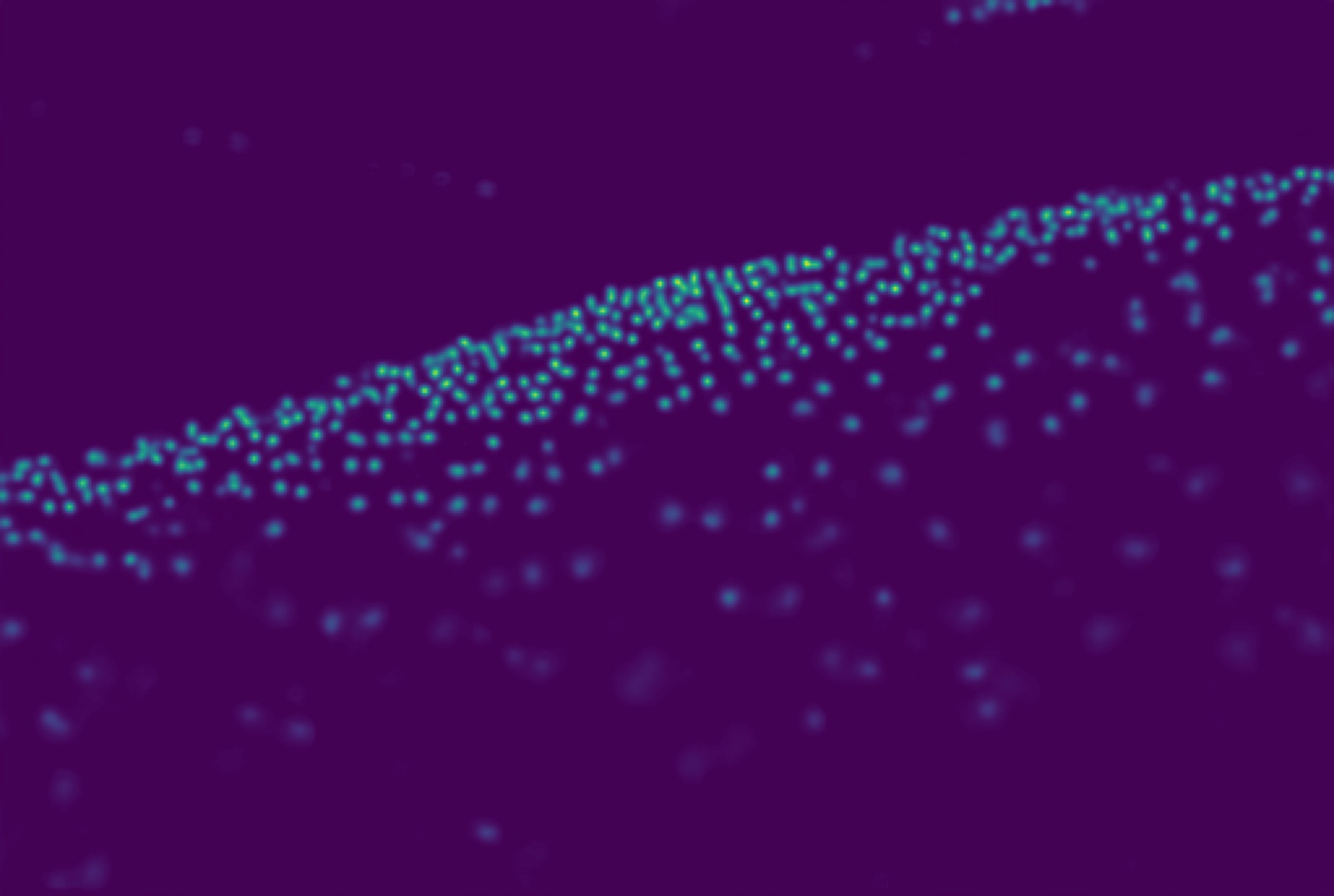}&
		\includegraphics[width=0.2\textwidth]{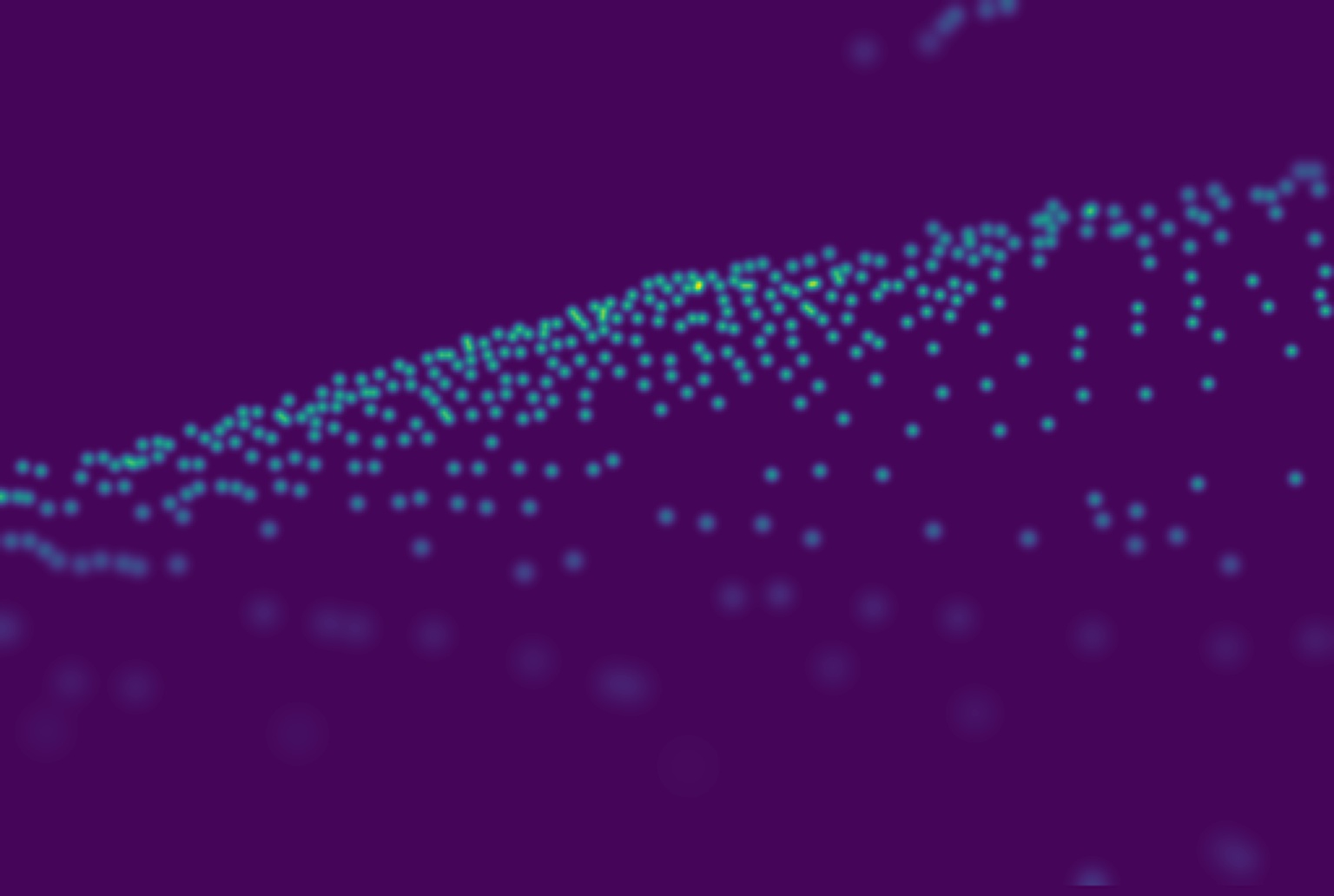}\\
		Estimated count & 404.8 & 466.1 & 389.7 & 381
	\end{tabular}
	\caption{Illustration of comparison examples against two of the state-of-the-art methods \cite{Li_2018_CVPR_CSRNet,Ma_2019_ICCV_BL}. }
	\label{fig:comparison}
\end{figure*}

\subsection{Comparison results}

In this section, we illustrate the quantitative results on different benchmarks including ShanghaiTech Part A and Part B~\cite{Zhang_2016_CVPR_MCNN}, UCF-QNRF~\cite{Idrees_2018_ECCV_CL}, and UCSD~\cite{chan2008privacy}. We demonstrate that the counting performance achieved by our simple network architecture can be better or comparable to the state-of-the-art models.

\textbf{ShanghaiTech Part A/B.}
To train our model, we follow the general training protocal. When training the initial counting model, we train the model for 100 epochs, and the learning rate is decayed by 10 times after training for 60, 80, and 90 epochs, respectively. During distillation, the learning rate will be reduced by 10 times after training for 20, 60, and 80 epochs.
We compared our method with some recent works, including MCNN~\cite{Zhang_2016_CVPR_MCNN}, Switching CNN~\cite{8099912_switch}, SANet~\cite{Cao_2018_ECCV_SA}, CSRNet~\cite{Li_2018_CVPR_CSRNet}, ic-CNN~\cite{ranjan2018iterative}, PACNN~\cite{Shi_2019_CVPR_PACNN}, ADCrowdNet~\cite{Liu_2019_CVPR_ADCrowd}, CAN~\cite{Liu_2019_CVPR_CAN}, BL~\cite{Ma_2019_ICCV_BL}, TEDnet~\cite{jiang2019crowd}, HA-CCN~\cite{sindagi2019ha}. The results are shown in Table~\ref*{Tab:Shanghaitech}. Although our model is based on a vanilla CNN architecture, our performance is comparable to the state-of-the-art works. For reference, our network architecture is similar to that of CSRNet~\cite{Li_2018_CVPR_CSRNet}, except that our network does not incorporate dilation convolution layers. As observed, depite our simple architecture, our result is significantly better than that of CSRNet.


\textbf{UCF-QNRF:}
To train and valiate our model in this dataset, we scale the long edge of each crowd image to 1080 pixels, while maintaining the origin aspect ratio. For data augmentation, we crop 4 patches from each image and each patch is 1/4 of the original image size. The learning rate is set as $10^{-6}$ and we decay the learning rate by 10 times at the $630^{th}$, $650^{th}$, $660^{th}$ epoch, respectively. During the distillation stage, the learning rate is set to $10^{-6}$ as well, and the learning rate will decay by 10 times after 80, 100, 110 epochs, respectively.
The comparison results of our method against the state-of-the-art methods, including MCNN~\cite{Zhang_2016_CVPR_MCNN}, MCNN~\cite{Zhang_2016_CVPR_MCNN}, Switching CNN~\cite{8099912_switch}, CL~\cite{Idrees_2018_ECCV_CL}, HA-CCN~\cite{sindagi2019ha}, RANet~\cite{zhang2019relational}, CAN~\cite{Liu_2019_CVPR_CAN}, TEDnet~\cite{jiang2019crowd}, SPN+L2SM~\cite{xu2019learn}, S-DCNet~\cite{xiong2019open}, SFCN~\cite{wang2019learning}, DSSINet~\cite{Liu_2019_ICCV_DSSINet}, MBTTBF-SCFB~\cite{Sindagi_2019_ICCV_MBTTBF}, are shown in Table~\ref{Tab:UCF-QNRF}. As observed, in the term of MAE, our model is demonstrated superior to the latest methods (e.g.~\cite{Sindagi_2019_ICCV_MBTTBF}) for at least more than $4.7\%$ gain.

\textbf{UCSD:}
During training, we upscale the images by two times and we leverage the provided regions of interest.
We first set the Gaussian $\sigma$ as 7 to generate the ground-truth density maps, and the set Gaussian $\sigma$ as 5 and 3 to produce density maps for distillation, respectively. Other settings follow the general training protocal.
As depicted in Table~\ref{Tab:UCSD}, compared with MCNN~\cite{Zhang_2016_CVPR_MCNN}, Switching CNN~\cite{8099912_switch}, ConvLSTM~\cite{xiong2017spatiotemporal}, BSAD~\cite{huang2017body}, ACSCP~\cite{shen2018crowd}, CSRNet~\cite{Li_2018_CVPR_CSRNet}, SANet~\cite{Cao_2018_ECCV_SA}, SPANet~\cite{Cheng_2019_ICCV_SPANet}, ADCrowdNet~\cite{Liu_2019_CVPR_ADCrowd}, our model shows the state-of-the-art performance. Since each image of this dataset contains a small number of people, the difference amongst the comparison methods are not significant. Even so, our approach can still be comparable to the latest methods (e.g.~\cite{Liu_2019_CVPR_ADCrowd}).

\textbf{Qualitative results:} In Fig.~\ref{fig:comparison}, we show several examples in which our results are compared against those of two representative crowd counting approaches, Bayesian loss based method \cite{Ma_2019_ICCV_BL} and CSRNet \cite{Li_2018_CVPR_CSRNet}. Perceptually, since our model generates the density maps with a relatively higher resolution, the visual quality of our produced density maps is much better and similar to the ground-truth.

\begin{figure*}
	\centering
	\setlength{\tabcolsep}{2pt}
	\begin{tabular}{cccccc}
		Input image & $G^{(t=0)}(25)$ & $G^{(t=1)}(20)$ & $G^{(t=2)}(10)$ & $G^{(t=3)}(5)$ & Ground-truth\\
		\includegraphics[width=0.16\textwidth]{images/dismap/IMG_72.jpg}&
		\includegraphics[width=0.16\textwidth]{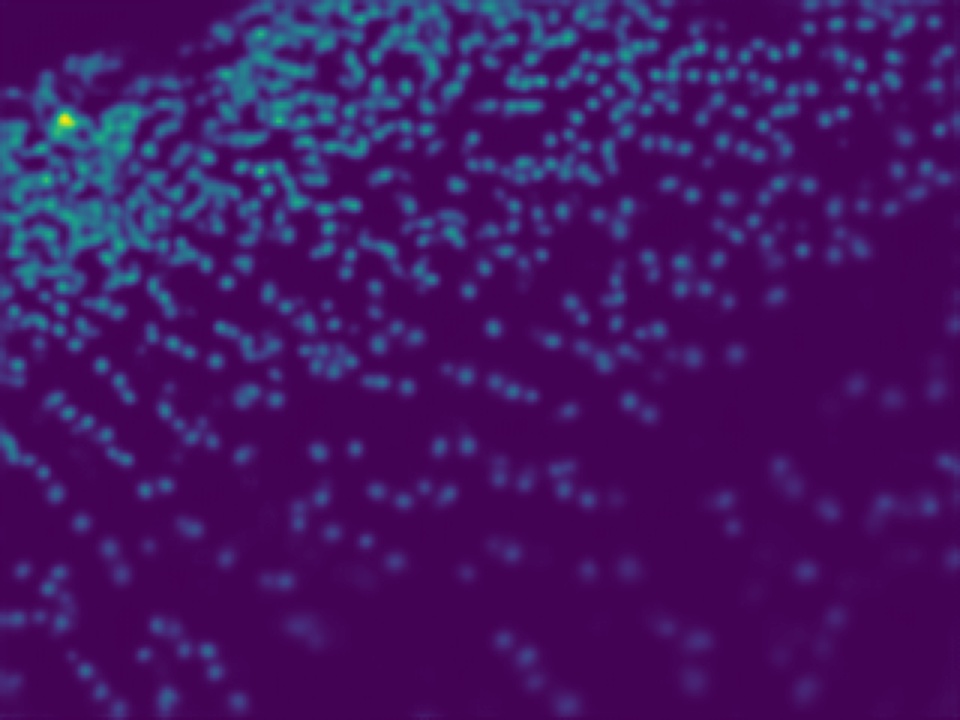}&
		\includegraphics[width=0.16\textwidth]{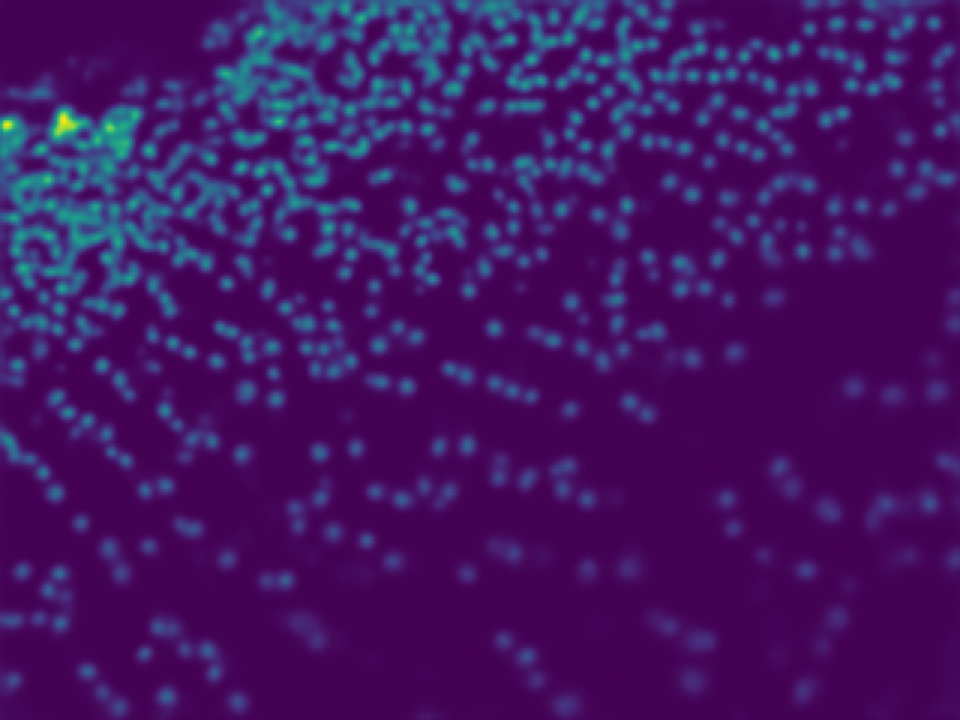}&
		\includegraphics[width=0.16\textwidth]{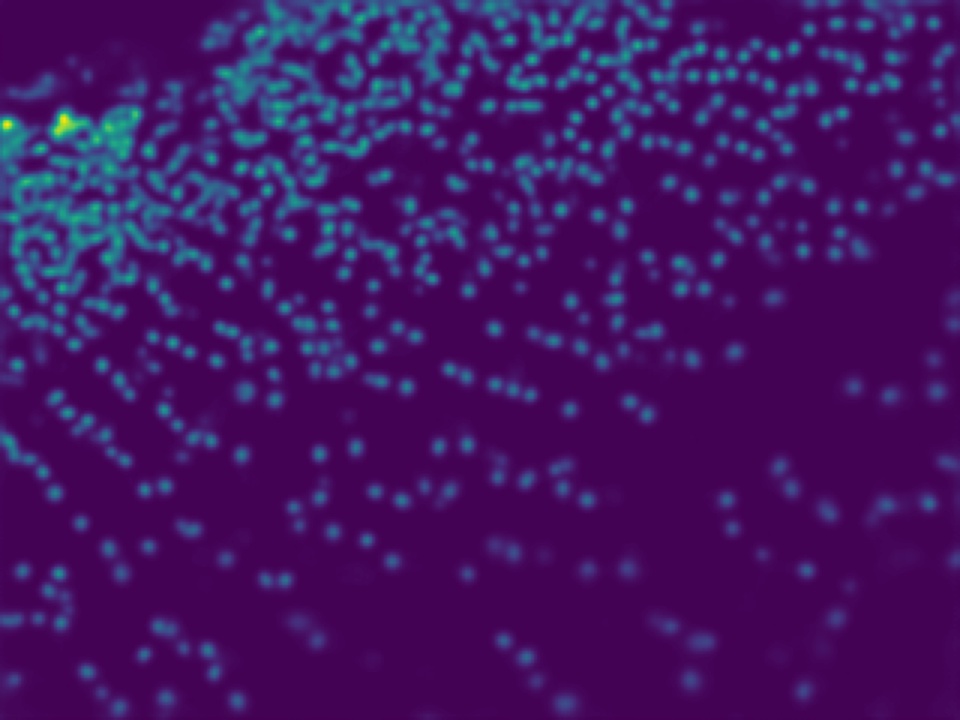}&
		\includegraphics[width=0.16\textwidth]{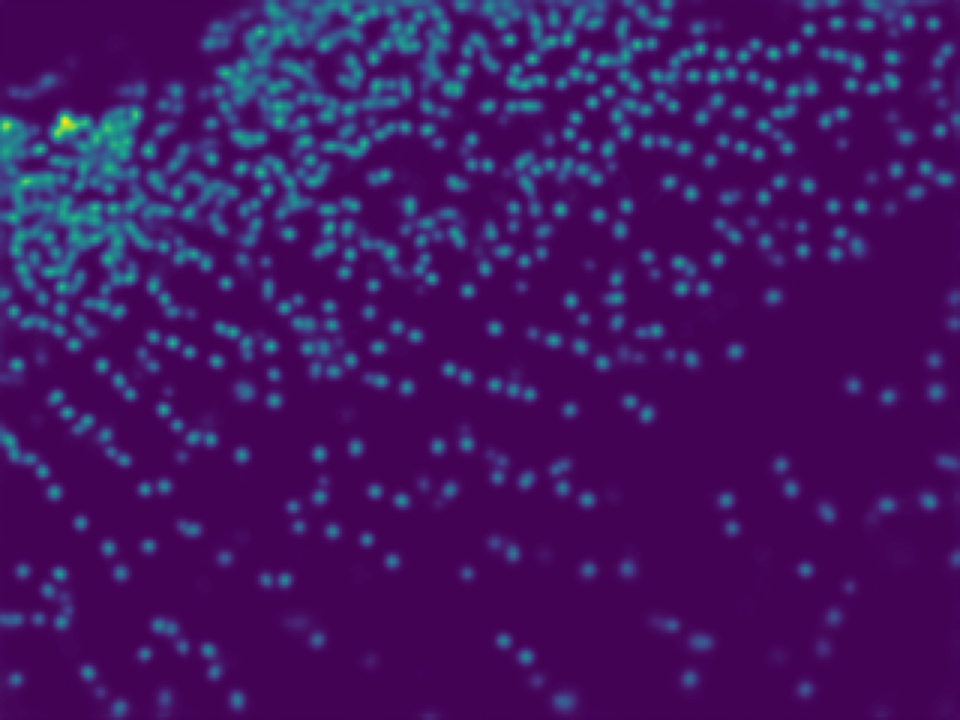}&
		\includegraphics[width=0.16\textwidth]{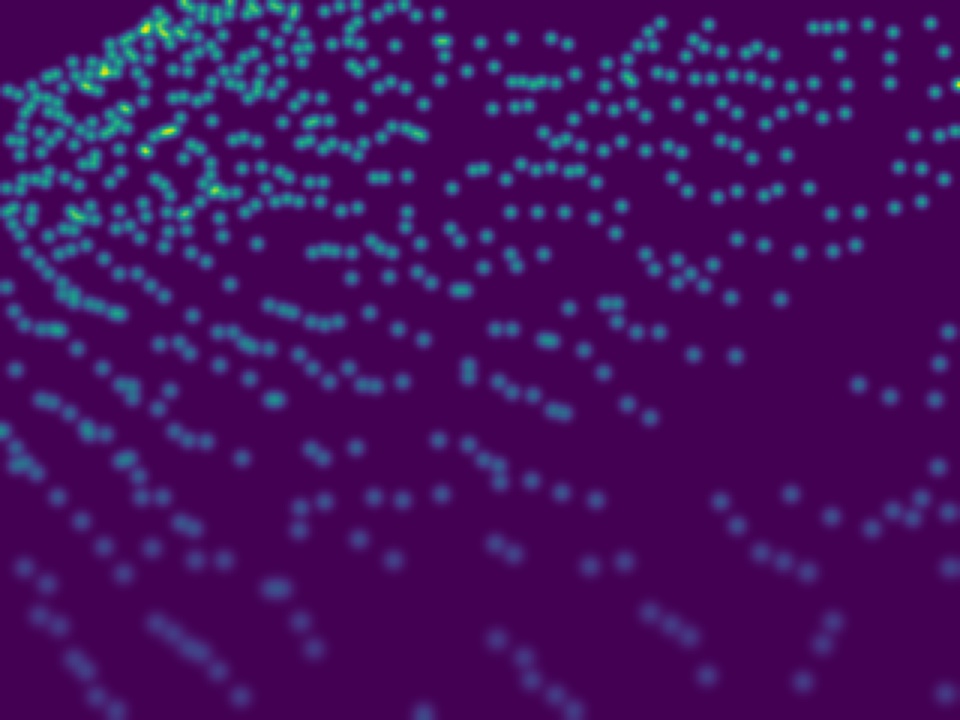}\\
		Estimated count & 697 & 686 & 678 & 679 & 589
	\end{tabular}
	\caption{Illustration of an example that progressively enhances the counting performance via multi-step distillation. }
	\label{fig:dif_map}
\end{figure*}

\subsection{Ablation study}
We perform the ablation study on the dataset, ShanghaiTech Part A~\cite{Zhang_2016_CVPR_MCNN}. 
we thoroughly dissect and delve into the structure of our method, including our generated ground-truth density maps, the multi-task loss of our applied network, and our distillation method.



\begin{figure*}
	\centering
	\includegraphics[width=1\linewidth]{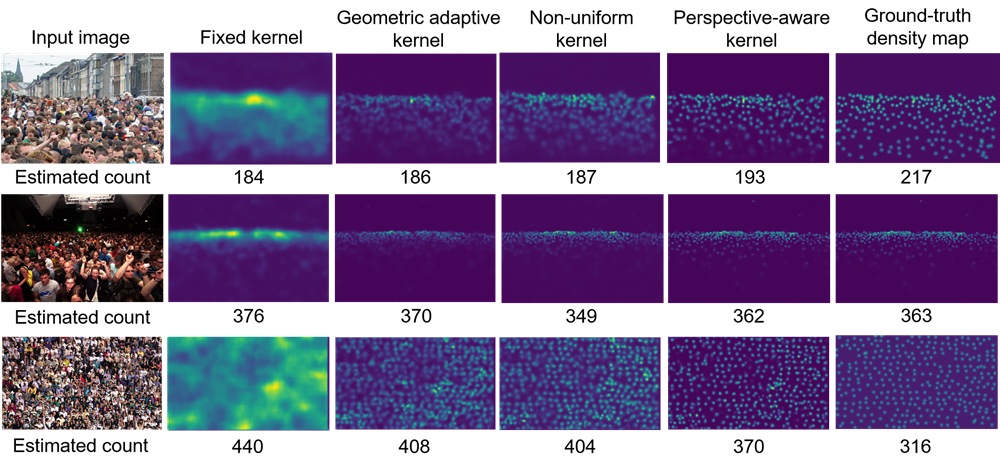}
	\caption{Illustration of examples generated by different density map generation methods. }
	\label{fig:dif_kernel}
\end{figure*}

\begin{table}
	\setlength{\tabcolsep}{4pt}
	\centering
	\caption{Ablation study on density map generation, multi-task loss, and distillation.
	}
	\begin{tabular}{c|c||c|c}
		\toprule
		Density map generation & MAE& Distillation & MAE \\
		\midrule
		+ Fixed & 67.91 & + $G^{(t=0)}(25)$ & 62.55 \\
		+ Adaptive & 66.25     
		&+ $G^{(t=1)}(20)$       &61.15\\
		+ Non-uniform & 64.58 &  + $G^{(t=2)}(10)$       &\textbf{61.09} \\
		+ Ours & \textbf{62.55}  &+ $G^{(t=3)}(5)$       &61.79\\
		\cline{3-4}
		& & + $G^{(t=0)}(10)$ & 62.72\\
		\midrule
		Single loss & 68.86 & Multi-task loss & 66.25\\
		\bottomrule
	\end{tabular}
	
	\label{tab:ablation}	
\end{table}

\textbf{Density map generation.} We apply different strategies, including fixed kernel size (``+Fixed"), geometric adaptive kernel size (``+Adaptive"), non-uniform kernel estimation (``+Non-uniform"), and our proposed perspective-aware kernel estimation, to generate ``ground-truth" density maps from point annotations to supervise our model. For comparison, the crowd counting network model is trained without distillation. 
The results are depicted in the left part of Table~\ref{tab:ablation}. As observed, our generated density maps can benefit the performance of crowd counting network, which significantly outperforms others by at least 2 points in the term of MAE.
In Fig.~\ref{fig:dif_kernel}, we illustrate several examples to show the difference of our proposed density map generator against previous methods. As observed, with the default settings, our approach can produce sharper density map than prior methods, which is able to deliver more accurate spatial information to the counting model and thus leads to better estimated counts.

\textbf{Distillation.} In the right part of Table~\ref{tab:ablation}, we iteratively run our distillation algorithm and measure the performance of the model on each stage. As observed, the distillation progressively promotes the counting performance of the model from 62.55 to 61.09, which is comparable to the state-of-the-art methods. But, when the kernel size shrinks to 5, the performance cannot further improve, due to the limit of distillation. 
After each distillation, the model will learn new supervision information on the original basis, optimize the performance of the model, and as the Gaussian kernel decreases, the density map will learn to have sharper position information. For reference, we also apply the density map generated by $G^{(t=0)}(10)$ to train our model from scratch. As observed, its result is worse than our model trained from $G^{(t=0)}(25)$. 

We illustrate an example of distillation in Fig.~\ref{fig:dif_map}. Particularly, we observe that, after distilling the model for two time steps (i.e., from $G^{(t=0)}$ to $G^{(t=2)}$), the results of the model are obviously improved, which may be due to the fact that the kernels of the density distribution near the camera becomes smaller and clearer. After distilling our model for more than 2 time steps, the performance hardly gains further improvements.

\textbf{Multi-task loss.} In the bottom row of Table~\ref{tab:ablation}, our baseline is a vanilla CNN-based model without the extra branch for supervising low-resolution density map (i.e., Single loss). Both networks are trained using the density maps produced by the geometric-adaptive kernel. As observed, the performance in the term of MAE (i.e., 68.86) is significantly worse than the one with multi-task loss (i.e., 66.25). 

\section{Conclusion and Future Works}
In this paper, we propose a perspective-aware density map generation method that is able to adaptively produce ground-truth density maps from point annotations to train crowd counting model to accomplish superior performance than prior density map generation techniques. Besides, leveraging our density map generation method, we propose an iterative distillation algorithm to progressively enhance our model with identical network structures, without significantly sacrificing the dimension of the output density maps. In experiments, we demonstrate that, with our simple convolutional neural network architecture strengthened by our proposed training algorithm, our model is able to outperform or be comparable with the state-of-the-art methods. 

Although our model can obtain satisfactory performance with a simple architecture, we have not validated our approach can be adapted to more complex network models. As the future work, we will explore the possibility of transferring our algorithm to advanced network models. On the other hand, distilling network is time consuming and there are many hyper-parameters for tuning. We will investigate more efficient algorithm to accomplish this purpose.

\ifCLASSOPTIONcaptionsoff
  \newpage
\fi

\bibliographystyle{IEEEtran}
\bibliography{IEEEabrv,IEEEexample}

\begin{thebibliography}{10}
\providecommand{\url}[1]{#1}
\csname url@samestyle\endcsname
\providecommand{\newblock}{\relax}
\providecommand{\bibinfo}[2]{#2}
\providecommand{\BIBentrySTDinterwordspacing}{\spaceskip=0pt\relax}
\providecommand{\BIBentryALTinterwordstretchfactor}{4}
\providecommand{\BIBentryALTinterwordspacing}{\spaceskip=\fontdimen2\font plus
\BIBentryALTinterwordstretchfactor\fontdimen3\font minus
  \fontdimen4\font\relax}
\providecommand{\BIBforeignlanguage}[2]{{%
\expandafter\ifx\csname l@#1\endcsname\relax
\typeout{** WARNING: IEEEtran.bst: No hyphenation pattern has been}%
\typeout{** loaded for the language `#1'. Using the pattern for}%
\typeout{** the default language instead.}%
\else
\language=\csname l@#1\endcsname
\fi
#2}}
\providecommand{\BIBdecl}{\relax}
\BIBdecl

\bibitem{Zhang_2016_CVPR_MCNN}
Y.~Zhang, D.~Zhou, S.~Chen, S.~Gao, and Y.~Ma, ``Single-image crowd counting
  via multi-column convolutional neural network,'' in \emph{The IEEE Conference
  on Computer Vision and Pattern Recognition (CVPR)}, June 2016.

\bibitem{Idrees_2018_ECCV_CL}
H.~Idrees, M.~Tayyab, K.~Athrey, D.~Zhang, S.~Al-Maadeed, N.~Rajpoot, and
  M.~Shah, ``Composition loss for counting, density map estimation and
  localization in dense crowds,'' in \emph{The European Conference on Computer
  Vision (ECCV)}, September 2018.

\bibitem{Ma_2019_ICCV_BL}
Z.~Ma, X.~Wei, X.~Hong, and Y.~Gong, ``Bayesian loss for crowd count estimation
  with point supervision,'' in \emph{The IEEE International Conference on
  Computer Vision (ICCV)}, October 2019.

\bibitem{Li_2018_CVPR_CSRNet}
Y.~Li, X.~Zhang, and D.~Chen, ``Csrnet: Dilated convolutional neural networks
  for understanding the highly congested scenes,'' in \emph{The IEEE Conference
  on Computer Vision and Pattern Recognition (CVPR)}, June 2018.

\bibitem{Yan_2019_ICCV_PGCNet}
Z.~Yan, Y.~Yuan, W.~Zuo, X.~Tan, Y.~Wang, S.~Wen, and E.~Ding,
  ``Perspective-guided convolution networks for crowd counting,'' in \emph{The
  IEEE International Conference on Computer Vision (ICCV)}, October 2019.

\bibitem{Liu_2019_ICCV_DSSINet}
L.~Liu, Z.~Qiu, G.~Li, S.~Liu, W.~Ouyang, and L.~Lin, ``Crowd counting with
  deep structured scale integration network,'' in \emph{The IEEE International
  Conference on Computer Vision (ICCV)}, October 2019.

\bibitem{Cao_2018_ECCV_SA}
X.~Cao, Z.~Wang, Y.~Zhao, and F.~Su, ``Scale aggregation network for accurate
  and efficient crowd counting,'' in \emph{The European Conference on Computer
  Vision (ECCV)}, September 2018.

\bibitem{Wan_2019_ICCV_adaptive_map}
J.~Wan and A.~Chan, ``Adaptive density map generation for crowd counting,'' in
  \emph{The IEEE International Conference on Computer Vision (ICCV)}, October
  2019.

\bibitem{NIPS2010_4043_map}
\BIBentryALTinterwordspacing
V.~Lempitsky and A.~Zisserman, ``Learning to count objects in images,'' in
  \emph{Advances in Neural Information Processing Systems 23}, J.~D. Lafferty,
  C.~K.~I. Williams, J.~Shawe-Taylor, R.~S. Zemel, and A.~Culotta, Eds.\hskip
  1em plus 0.5em minus 0.4em\relax Curran Associates, Inc., 2010, pp.
  1324--1332. [Online]. Available:
  \url{http://papers.nips.cc/paper/4043-learning-to-count-objects-in-images.pdf}
\BIBentrySTDinterwordspacing

\bibitem{zhang2015cross}
C.~Zhang, H.~Li, X.~Wang, and X.~Yang, ``Cross-scene crowd counting via deep
  convolutional neural networks,'' in \emph{Proceedings of the IEEE conference
  on computer vision and pattern recognition}, 2015, pp. 833--841.

\bibitem{hinton2015distilling}
G.~Hinton, O.~Vinyals, and J.~Dean, ``Distilling the knowledge in a neural
  network,'' \emph{arXiv preprint arXiv:1503.02531}, 2015.

\bibitem{yang2019snapshot}
C.~Yang, L.~Xie, C.~Su, and A.~L. Yuille, ``Snapshot distillation:
  Teacher-student optimization in one generation,'' in \emph{Proceedings of the
  IEEE Conference on Computer Vision and Pattern Recognition}, 2019, pp.
  2859--2868.

\bibitem{furlanello2018born}
T.~Furlanello, Z.~C. Lipton, M.~Tschannen, L.~Itti, and A.~Anandkumar, ``Born
  again neural networks,'' \emph{arXiv preprint arXiv:1805.04770}, 2018.

\bibitem{shi2019counting}
Z.~Shi, P.~Mettes, and C.~G. Snoek, ``Counting with focus for free,'' in
  \emph{Proceedings of the IEEE International Conference on Computer Vision},
  2019, pp. 4200--4209.

\bibitem{simonyan2014very}
K.~Simonyan and A.~Zisserman, ``Very deep convolutional networks for
  large-scale image recognition,'' \emph{arXiv preprint arXiv:1409.1556}, 2014.

\bibitem{Sindagi_2019_ICCV_MBTTBF}
V.~A. Sindagi and V.~M. Patel, ``Multi-level bottom-top and top-bottom feature
  fusion for crowd counting,'' in \emph{The IEEE International Conference on
  Computer Vision (ICCV)}, October 2019.

\bibitem{Liu_2019_CVPR_ADCrowd}
N.~Liu, Y.~Long, C.~Zou, Q.~Niu, L.~Pan, and H.~Wu, ``Adcrowdnet: An
  attention-injective deformable convolutional network for crowd
  understanding,'' in \emph{The IEEE Conference on Computer Vision and Pattern
  Recognition (CVPR)}, June 2019.

\bibitem{Shi_2019_CVPR_PACNN}
M.~Shi, Z.~Yang, C.~Xu, and Q.~Chen, ``Revisiting perspective information for
  efficient crowd counting,'' in \emph{The IEEE Conference on Computer Vision
  and Pattern Recognition (CVPR)}, June 2019.

\bibitem{Liu_2019_CVPR_CAN}
W.~Liu, M.~Salzmann, and P.~Fua, ``Context-aware crowd counting,'' in \emph{The
  IEEE Conference on Computer Vision and Pattern Recognition (CVPR)}, June
  2019.

\bibitem{8099912_switch}
D.~B. {Sam}, S.~{Surya}, and R.~V. {Babu}, ``Switching convolutional neural
  network for crowd counting,'' in \emph{2017 IEEE Conference on Computer
  Vision and Pattern Recognition (CVPR)}, July 2017, pp. 4031--4039.

\bibitem{xu2019learn}
C.~Xu, K.~Qiu, J.~Fu, S.~Bai, Y.~Xu, and X.~Bai, ``Learn to scale: Generating
  multipolar normalized density maps for crowd counting,'' in \emph{Proceedings
  of the IEEE International Conference on Computer Vision}, 2019, pp.
  8382--8390.

\bibitem{xiong2019open}
H.~Xiong, H.~Lu, C.~Liu, L.~Liu, Z.~Cao, and C.~Shen, ``From open set to closed
  set: Counting objects by spatial divide-and-conquer,'' in \emph{Proceedings
  of the IEEE International Conference on Computer Vision}, 2019, pp.
  8362--8371.

\bibitem{shen2018crowd}
Z.~Shen, Y.~Xu, B.~Ni, M.~Wang, J.~Hu, and X.~Yang, ``Crowd counting via
  adversarial cross-scale consistency pursuit,'' in \emph{Proceedings of the
  IEEE conference on computer vision and pattern recognition}, 2018, pp.
  5245--5254.

\bibitem{chan2008privacy}
A.~B. Chan, Z.-S.~J. Liang, and N.~Vasconcelos, ``Privacy preserving crowd
  monitoring: Counting people without people models or tracking,'' in
  \emph{2008 IEEE Conference on Computer Vision and Pattern Recognition}.\hskip
  1em plus 0.5em minus 0.4em\relax IEEE, 2008, pp. 1--7.

\bibitem{ranjan2018iterative}
V.~Ranjan, H.~Le, and M.~Hoai, ``Iterative crowd counting,'' in
  \emph{Proceedings of the European Conference on Computer Vision (ECCV)},
  2018, pp. 270--285.

\bibitem{jiang2019crowd}
X.~Jiang, Z.~Xiao, B.~Zhang, X.~Zhen, X.~Cao, D.~Doermann, and L.~Shao, ``Crowd
  counting and density estimation by trellis encoder-decoder networks,'' in
  \emph{Proceedings of the IEEE Conference on Computer Vision and Pattern
  Recognition}, 2019, pp. 6133--6142.

\bibitem{sindagi2019ha}
V.~A. Sindagi and V.~M. Patel, ``Ha-ccn: Hierarchical attention-based crowd
  counting network,'' \emph{IEEE Transactions on Image Processing}, vol.~29,
  pp. 323--335, 2019.

\bibitem{zhang2019relational}
A.~Zhang, J.~Shen, Z.~Xiao, F.~Zhu, X.~Zhen, X.~Cao, and L.~Shao, ``Relational
  attention network for crowd counting,'' in \emph{Proceedings of the IEEE
  International Conference on Computer Vision}, 2019, pp. 6788--6797.

\bibitem{wang2019learning}
Q.~Wang, J.~Gao, W.~Lin, and Y.~Yuan, ``Learning from synthetic data for crowd
  counting in the wild,'' in \emph{Proceedings of the IEEE conference on
  computer vision and pattern recognition}, 2019, pp. 8198--8207.

\bibitem{xiong2017spatiotemporal}
F.~Xiong, X.~Shi, and D.-Y. Yeung, ``Spatiotemporal modeling for crowd counting
  in videos,'' in \emph{Proceedings of the IEEE International Conference on
  Computer Vision}, 2017, pp. 5151--5159.

\bibitem{huang2017body}
S.~Huang, X.~Li, Z.~Zhang, F.~Wu, S.~Gao, R.~Ji, and J.~Han, ``Body structure
  aware deep crowd counting,'' \emph{IEEE Transactions on Image Processing},
  vol.~27, no.~3, pp. 1049--1059, 2017.

\bibitem{Cheng_2019_ICCV_SPANet}
Z.-Q. Cheng, J.-X. Li, Q.~Dai, X.~Wu, and A.~G. Hauptmann, ``Learning spatial
  awareness to improve crowd counting,'' in \emph{The IEEE International
  Conference on Computer Vision (ICCV)}, October 2019.

\end{thebibliography}



%




\end{document}